\documentclass[letterpaper, 10 pt, conference]{ieeeconf}  %
\usepackage{FG2025}

\usepackage{graphicx}
\usepackage{amsmath}
\usepackage{amssymb}
\usepackage{booktabs}
\usepackage[pagebackref,breaklinks,colorlinks]{hyperref}
\usepackage[capitalize]{cleveref}
\usepackage{colortbl}
\usepackage{multirow}
\usepackage{subcaption}

\FGfinalcopy %

\newcommand{\tablefontsize}[1]{%
  \fontsize{#1}{#1}\selectfont
}

\IEEEoverridecommandlockouts                              %
\def\FGPaperID{84} %

\title{\LARGE \bf
Impact of Sunglasses on One-to-Many Facial Identification Accuracy}

\author{\parbox{16cm}{\centering
    {\large Sicong Tian$^1$, Haiyu Wu$^2$, Michael C. King$^3$, Kevin W. Bowyer$^2$}\\
    {\normalsize
    $^1$Indiana University South Bend,
    $^2$University of Notre Dame,
    $^3$Florida Institute of Technology}}
}

\begin{document}

\ifFGfinal
\thispagestyle{empty}
\pagestyle{empty}
\else
\author{Anonymous FG2025 submission\\ Paper ID \FGPaperID \\}
\pagestyle{plain}
\fi
\maketitle

\begin{abstract}
One-to-many facial identification is documented to achieve high accuracy in the case where both the probe and the gallery are `mugshot quality' images. However, an increasing number of documented instances of wrongful arrest following one-to-many facial identification have raised questions about its accuracy. Probe images used in one-to-many facial identification are often cropped from frames of surveillance video and deviate from `mugshot quality' in various ways. 
This paper systematically explores how the accuracy of one-to-many facial identification is degraded by the person in the probe image choosing to wear dark sunglasses.
We show that sunglasses degrade accuracy for mugshot-quality images by an amount similar to strong blur or noticeably lower resolution. Further, we demonstrate that the combination of sunglasses with blur or lower resolution results in even more pronounced loss in accuracy. These results have important implications for developing objective criteria to qualify a probe image for the level of accuracy to be expected if it used for one-to-many identification.
To ameliorate the accuracy degradation caused by dark sunglasses, we show that it is possible to recover %
about 38\% of the lost accuracy by synthetically adding sunglasses to all the gallery images, 
without model re-training. We also show that the frequency of wearing-sunglasses images is very low in existing training sets, and that increasing the representation of wearing-sunglasses images can greatly reduce the error rate.
The image set assembled for this research is available at \url{https://cvrl.nd.edu/projects/data/} to support replication and further research. 

\end{abstract}

\section{Introduction}
\label{sec:intro}

One-to-many facial identification has attracted significant attention due to it being used in a growing number of cases of wrongful arrest~\cite{wrong-arrest-Murphy, wrong-arrest-WilliamsOliverParks, wrong-arrest-Reid, wrong-arrest-Woodruff}. 
In real-world scenarios, probe images used in one-to-many facial identification are often extracted from surveillance video. Such images may suffer from quality issues such as blur, low resolution, off-angle pose, and the presence of occlusions such as sunglasses and / or hats. Recent work has systematically investigated how image-based factors like resolution and blur degrade the accuracy of one-to-many identification~\cite{bhatta-one-to-many}. 
However, subject-based factors such as wearing sunglasses or caps, or extreme pitch angles, which may be common in surveillance camera images and may substantially degrade identification accuracy, have not been thoroughly addressed. 
\begin{figure}[t]
    \centering
        \includegraphics[width=\linewidth]{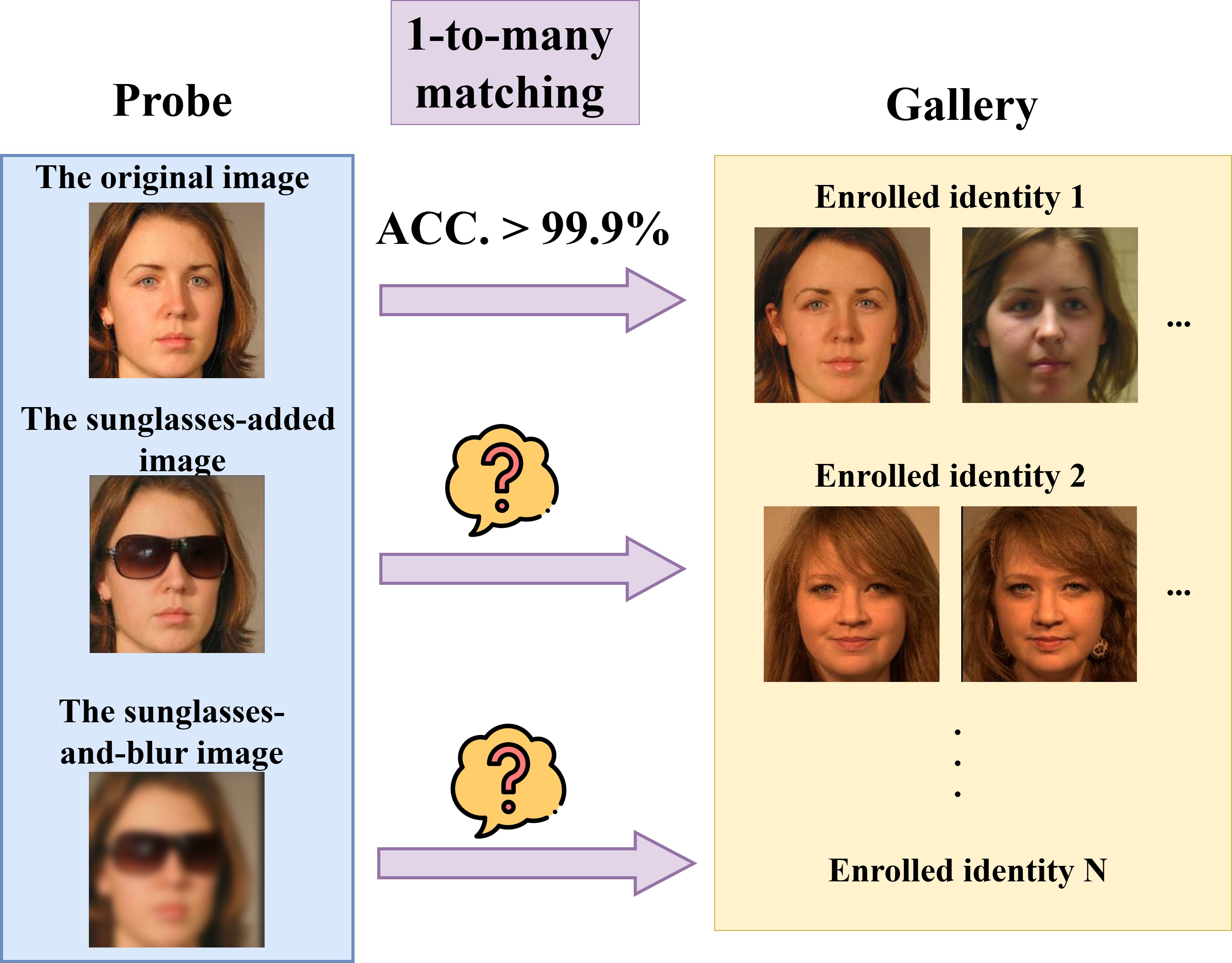}
   \caption{How much does wearing sunglasses degrade one-to-many facial identification accuracy? One-to-many identification has high accuracy when the probe and gallery images are all high quality \cite{Grother_NIST_2019}. This work explores how accuracy degrades when the person in the probe image wears sunglasses, and alternatives for ameliorating this problem.   
   }
   \vspace{-4mm}
\label{fig:images/teaser_figure}
\end{figure}

This paper presents the systematic exploration of the effect of sunglasses on the accuracy of one-to-many facial identification. First, to create a research dataset for this work, we assemble a test set, named ``ND-sunglasses", that contains 15,088 high-quality image pairs (original and with sunglasses), where the only difference in each pair of images is the sunglasses region. Then, we sample a range of degrees of blur and lower resolutions to select ones that have a similar-sized effect on accuracy as sunglasses. Then, we systematically analyze the impact of the combination of sunglasses, blur, and low resolution on accuracy. The results reveal a pronounced accuracy degradation, showing the importance of this study. Next, we introduce a simple but effective approach that can recover up to 38\% of the accuracy lost, without model re-training, by adding sunglasses to the gallery images. Lastly, we also
show that increasing the representation of wearing-sunglasses
images in the training set can greatly reduce the error rate.

Our study contributes to a better understanding of the accuracy that can be expected from one-to-many facial identification with probe images taken from surveillance video.
The accuracy is far lower than has been documented for mugshot-quality images \cite{Grother_NIST_2019, Grother_NIST_2019b, Krishnapriya_TTS_2020}, and a variety of different factors can contribute to degrading the accuracy.

Contributions of this work include the following: 
\begin{itemize}
    \item Experimental results showing that sunglasses in the probe image significantly reduce the accuracy of one-to-many identification.
\vspace{1mm}
    \item Analysis showing that sunglasses have an additive effect in combination with blur or reduced resolution of the face in the probe image, as may be typical of real-world surveillance camera images.
\vspace{1mm}
    \item Results showing that when a probe image contains sunglasses while the gallery images do not, adding sunglasses to the gallery images can mitigate the accuracy loss due to sunglasses in the probe.
\vspace{1mm}
    \item Increasing the representation of sunglasses-wearing images in the training dataset substantially reduces error rates. This finding offers a practical approach to mitigating the accuracy loss associated with sunglasses.
\vspace{1mm}
    \item Releasing the first  dataset focused on studying the effect of sunglasses in one-to-many facial identification to enable replication and extension of our results.
\end{itemize}

\section{Related Work}
\label{sec:Relate Work}
\begin{figure}[t]
    \centering
        \includegraphics[width=1.0\linewidth]{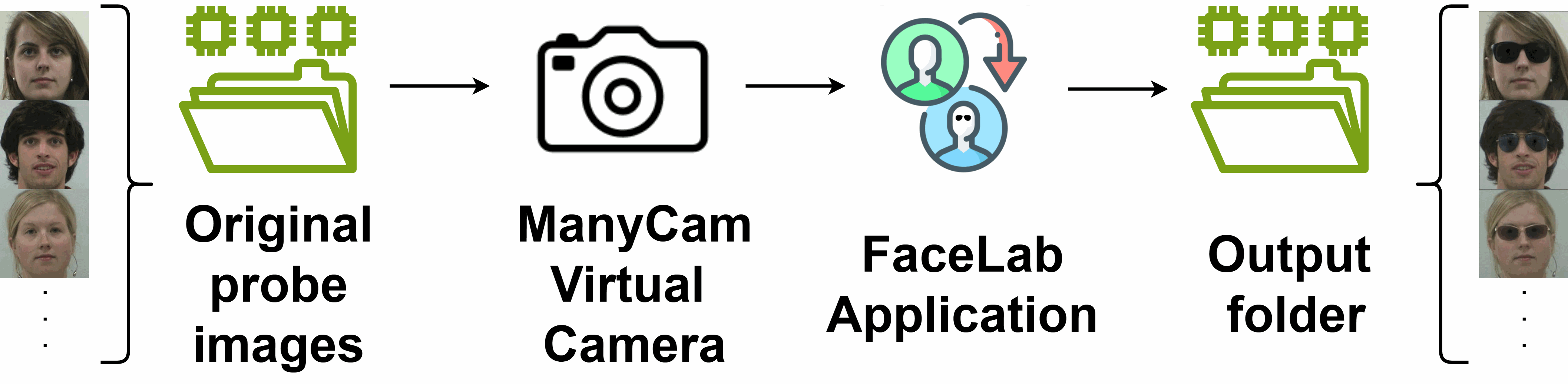}
   \vspace{-4mm}
   \caption{The pipeline for creating sunglasses-added versions of original images. An Android emulator and ManyCam are used to handle the image input and processing steps, the FaceLab app adds sunglasses, and a screenshot of the FaceLab results is stored.}
   \vspace{-4mm}
\label{fig:images/facelab_pipeline.drawio}
\end{figure}

The NIST report~\cite{Grother_NIST_2019} stands as the largest experimental investigation of the accuracy of one-to-many facial identification, given the large number of matchers evaluated and the size of image datasets used. The report emphasizes that demographic differences in the false positive identification rate (FPIR) are significantly larger than those in the false negative identification rate (FNIR), with FPIR differences potentially reaching one or two orders of magnitude (i.e., 10x, 100x). In contrast, FNIR differences are  smaller and more specific to the algorithms used.
Krishnapriya et al.\cite{Krishnapriya_TTS_2020} also examined one-to-many matching accuracy, utilizing the MORPH~\cite{morph} dataset. Their findings agree with the NIST study, that when the person in the probe  has an enrolled image, the one-to-many search is highly likely to return the correct identity. Both~\cite{Grother_NIST_2019} and~\cite{Krishnapriya_TTS_2020} are clear to state that, if the person in the probe image does not have an image enrolled in the gallery, any result returned by one-to-many facial identification must be a false positive identification.

Another NIST report~\cite{Grother_NIST_2019b} discusses the impressive accuracy achieved when matching mugshot-quality probe images to mugshot-quality enrolled images -- ``... in galleries containing 12 million individuals, with rank one miss rates approaching 0.1\%.'' 
(The term ``mugshot quality" refers generally to face images with frontal pose, neutral expression and controlled lighting and background, as also seen with driver's licence or visa / passport images.)
The NIST report notes that the primary datasets comprise good quality images, which is an important factor in the low FPIRs achieved by the algorithms.
However, this NIST report also notes that, ``Recognition in other circumstances, particularly those without a dedicated photographic environment and human or automated quality control checks, will lead to declines in accuracy.''

One factor in the documented instances of wrongful arrest following one-to-many facial identification \cite {wrong-arrest-Murphy, wrong-arrest-WilliamsOliverParks, wrong-arrest-Reid, wrong-arrest-Woodruff} appears to be a mistaken confidence in the accuracy that can be achieved when the probe image is surveillance camera quality rather than mugshot quality.
This has led to work to document how factors that can occur in face images from surveillance video degrade the accuracy of one-to-many facial identification.

Bhatta {\it et al.} \cite{bhatta-one-to-many} investigate how image quality factors such as blur and reduced resolution affect one-to-many search using the MORPH dataset. They report that increased blur or reduced resolution of the face in the probe image can significantly increase the FPIR. 
They also find a larger accuracy degradation for females than for males.

In addition to image-based factors such as blur and resolution, face images taken from surveillance video may capture a person wearing sunglasses.
Existing face attribute editing methods could potentially be used to add sunglasses to a face image \cite{attgan, stgan, meglass, eyeglasses-add}. However, these methods suffer from unsatisfactory performance on high-resolution images. Moreover, due to the lack of sunglasses labels in popular existing training sets for GANs, adding sunglasses to an image is not (yet) a stable operation with these approaches. Hence, we choose to use a commercial (non-GAN) tool to reliably add realistic-looking sunglasses without changing other regions of the image, so that the addition of sunglasses is identity-preserving.

One related work~\cite{sunglass-work} proposes an occlusion dataset for one-to-many research, but this dataset has only 161 probe identities and lacks gender and race labels to address potential ethical bias, preventing statistically meaningful analysis. Additionally, Wang and Kumar~\cite{wang2016recognizing} studied how disguises and makeup affect face recognition accuracy, demonstrating significant performance degradation with disguised faces. However, their work focused on general face recognition rather than one-to-many identification. 

This paper extends recent work by exploring the degradation in one-to-many facial identification accuracy caused by wearing sunglasses and providing two possible solutions to mitigate this degradation.

\section{Dataset}

\label{sec:Data Collection and Matcher}

\begin{figure}[t]
    \begin{subfigure}[b]{1\linewidth}
    \centering
        \includegraphics[width=0.9\linewidth]{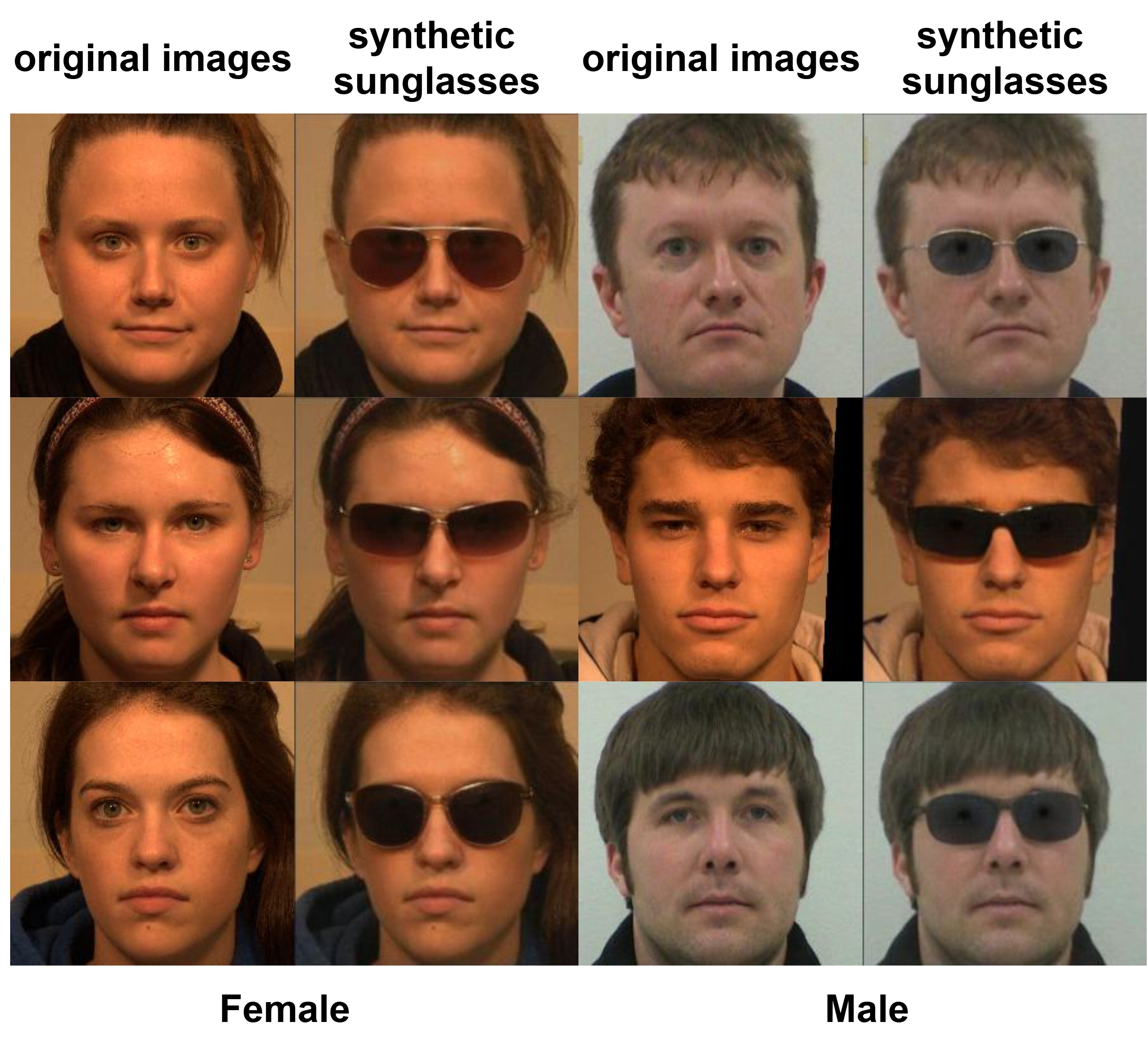}
    \end{subfigure}
    \vspace{-6mm}
   \caption{Examples of original images and synthetic versions with sunglasses added using FaceLab app. Adding synthetic sunglasses alters only the sunglasses region of the original image. Note that the style of the synthetic sunglasses varies between images.}
\vspace{-5mm}
\label{fig:facelab_example}
\end{figure}

There are specific requirements for a dataset suitable for exploring the effect of sunglasses in probe images on the accuracy of one-to-many identification.
One, similar quality images of the same persons, wearing sunglasses in one image and not in another, are required for use as probe images for comparison.
Two, additional images of the same persons are required for use as the enrolled gallery.
The gallery images of a person should be acquired in different sessions than the probe images, to avoid any ``same-session effect'' in accuracy estimates.
Three, images should be frontal to reduce the impact of head pose variation, allowing us to focus specifically on how sunglasses affect accuracy.

There are existing test sets of face images covering a variety of aspects of facial recognition, including pose \cite{cplfw,cfpfp}, age difference \cite{calfw, agedb-30}, size of face in image \cite{tinyface}, in-the-wild images and video \cite{lfw, ijba, ijbb, ijbc}, and illumination and facial hair \cite{hadrian-eclipse}.
None of these focus specifically on the effects of sunglasses.
The IJB datasets \cite{ijba, ijbb, ijbc} contain images of some persons wearing sunglasses.
However, the IJB datasets emphasize in-the-wild imagery, which would be appropriate for probe images in our experiments, but not gallery images.
Also, the presence of similar quality images of the same person, both wearing and not wearing sunglasses, would happen only by accident in the IJB datasets.
Additionally, NIST has discontinued distribution of the IJB datasets.\footnote{https://www.nist.gov/itl/iad/ig/ijb-c-dataset-request-form}
The AR face image dataset \cite{Martinez_AR_1998} introduced by Martinez contains images of persons wearing sunglasses.
However, the dataset only contains 126 people with grayscale images rather than color images, all subjects wear the same model/style of sunglasses, and since images were acquired in two sessions per person, each person could have only one gallery image in one-to-many identification experiments
In the end, we could not identify an existing dataset suitable for exploring the effects of wearing sunglasses in the context of one-to-many facial identification.

\begin{figure}[t]
    \centering
    \begin{subfigure}[b]{1\linewidth}
        \includegraphics[width=\linewidth]{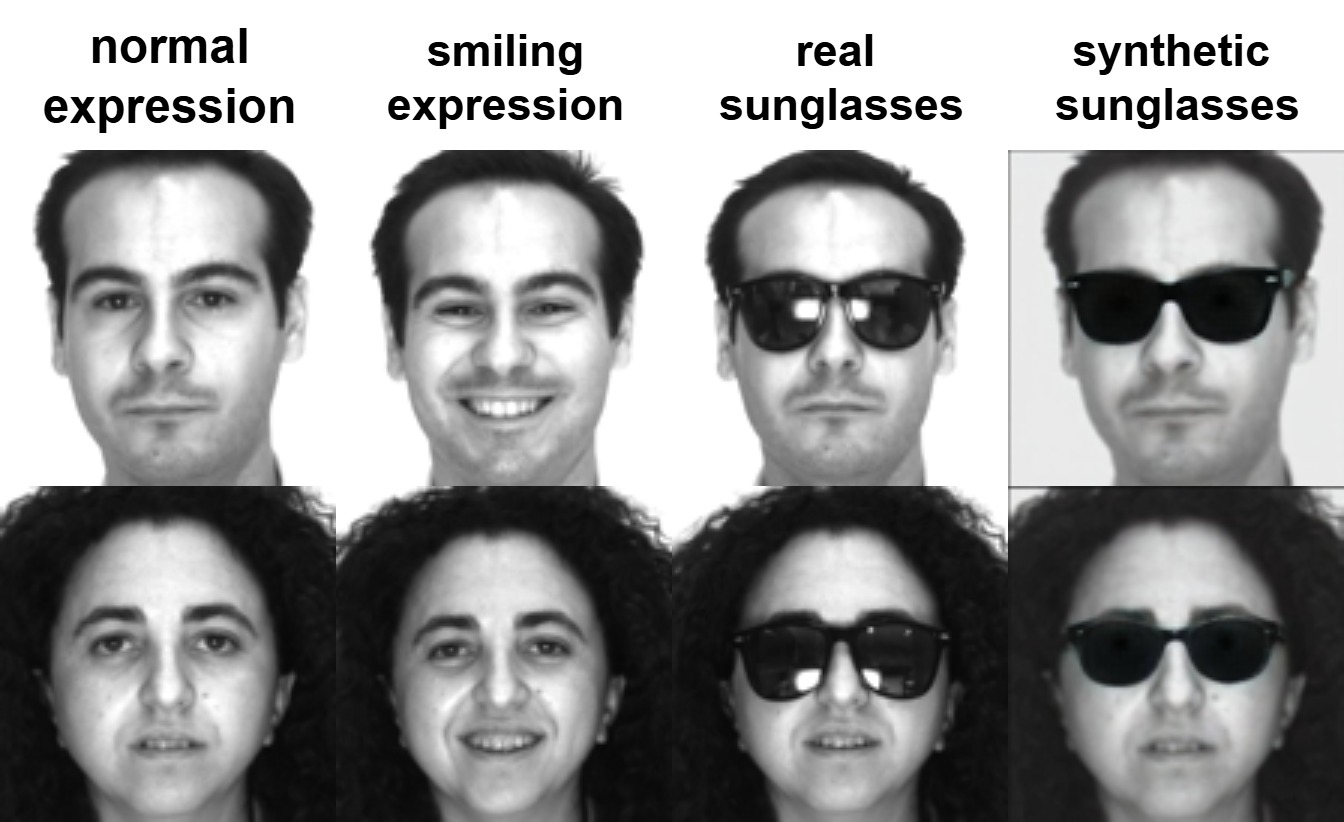}
    \end{subfigure}
    \vspace{-5mm}
   \caption{Examples of FaceLab's synthetic sunglasses on AR dataset. This 2x4 matrix displays two subjects (rows) under four conditions (columns): normal expression, smiling expression, real sunglasses, and synthetic sunglasses applied to the normal expression.}
\vspace{-4mm}
\label{fig:AR_example}
\end{figure}

To assemble a suitable dataset for one-to-many identification experiments, we started with the Notre Dame Male/Female Accuracy Dataset (ND-MFAD) \cite{ND_MFAD, Albiero_IJCB_2022, Bhatta_WACVW_2023, notre-dame}. This dataset has several advantages.  
Images were acquired in a controlled environment over a number of different sessions, so that a gallery of mugshot-quality images can be created.
Also, all subjects completed IRB-approved informed consent for their images to be used in research. 
For our experiments, the images were cropped and aligned using img2pose~\cite{img2pose}. The ND-MFAD dataset contains images of 575 subjects who identified as Caucasian female, with a total of 6,669 images, and of 687 subjects who identified as Caucasian male, with a total of 8,419 images.

\begin{figure}[t]
    \centering
    \begin{subfigure}[b]{1\linewidth}
        \includegraphics[width=\linewidth]{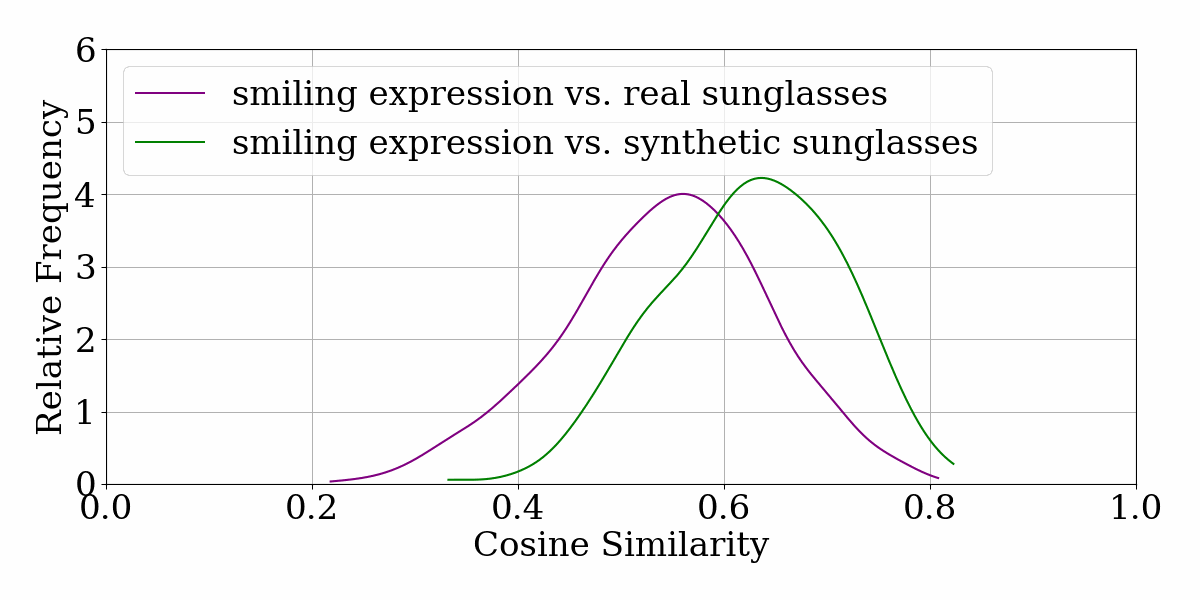}
    \end{subfigure}
    \vspace{-8mm}
   \caption{Cosine similarity distributions comparing images with smiling expressions to those with real and synthetic sunglasses. The observed shift towards higher similarity scores with synthetic sunglasses, compared to real ones, shows that the FaceLab app's sunglasses addition can preserve identity.}
\vspace{-4mm}
\label{fig:AR_distribution}
\end{figure}

\begin{figure}[h]
    \centering
        \includegraphics[width=0.8\linewidth]{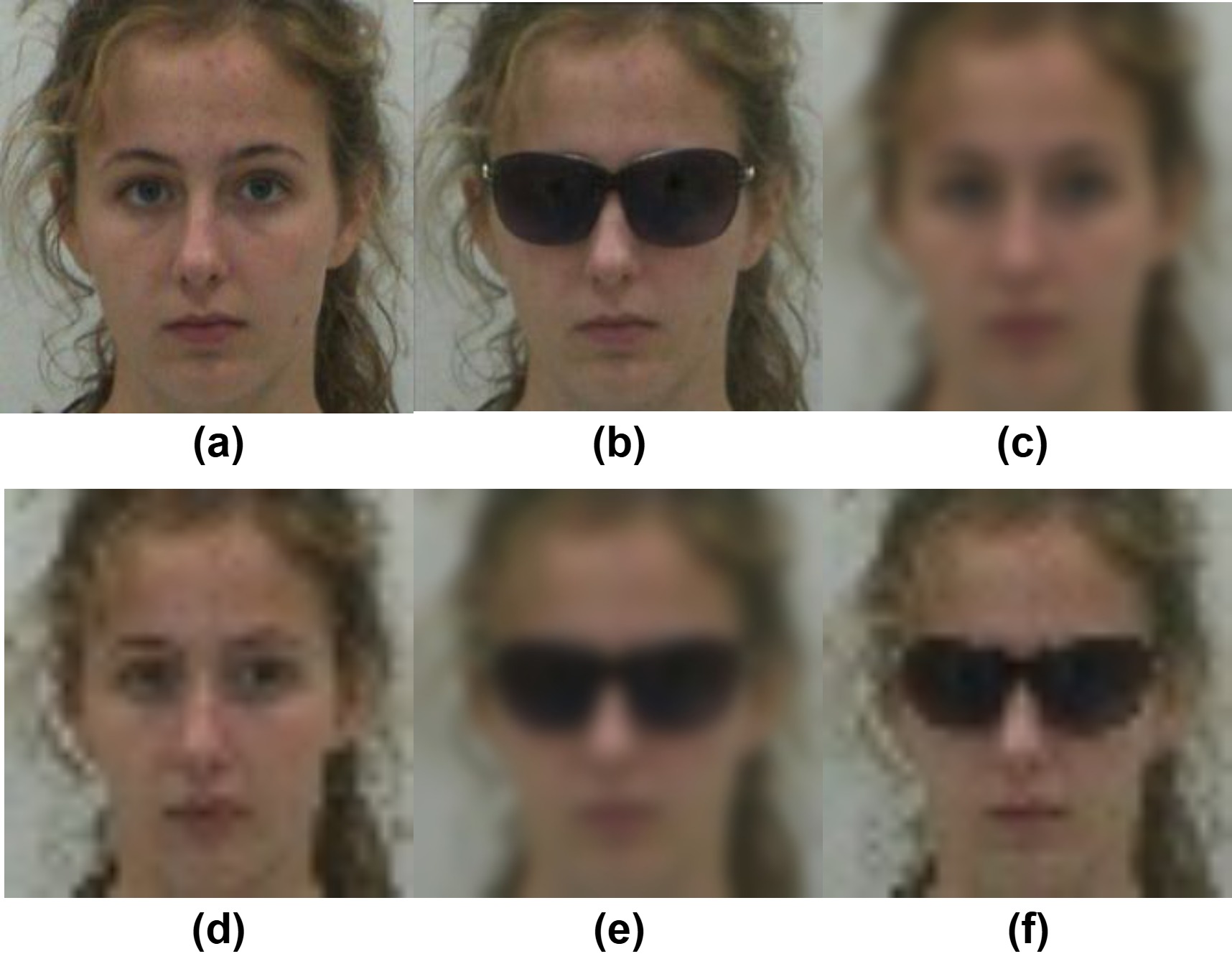}
    \vspace{-2mm}
   \caption{Example of probe images for six conditions considered in our work. (a) original image, (b) image with sunglasses, (c) blur ($\sigma=4.6$), that degrades accuracy about equal to sunglasses, (d) low resolution (37x37), that degrades accuracy about equal to sunglasses, (e) sunglasses + blur (f) sunglasses + low resolution.}
   \vspace{-5mm}
\label{fig:images/examples}
\end{figure}

In this study, we used a commercial app, ``FaceLab\footnote{https://lyrebirdstudio.net/},'' to add sunglasses to each original image. Figure~\ref{fig:images/facelab_pipeline.drawio} illustrates the pipeline of adding sunglasses to the images. Our experimental procedure involved processing 15,088 images through FaceLab to add sunglasses. 
Occasional issues, such as emulator crashes and processing failures, resulted in an initial 2\% to 3\% failure rate. These images were manually processed to have a ``sunglasses-added'' version for all the original images. Examples of images with synthetic sunglasses added by FaceLab are shown in Figure~\ref{fig:facelab_example}.

While the AR dataset is not appropriate for one-to-many identification experiments, it does contain a subset of images featuring subjects wearing real sunglasses. This characteristic allows us to evaluate the identity preservation capability of the FaceLab app when synthetically adding sunglasses to images. To conduct this assessment, we applied synthetic sunglasses to the normal expression images from the AR database using FaceLab. Examples of image comparisons are shown in Figure~\ref{fig:AR_example}. 
We then analyzed the cosine similarity distributions, shown in Figure~\ref{fig:AR_distribution}, 
between matching (a) image of a person smiling to image of the person wearing real sunglasses, and (b) same image of a person smiling to image of the person with synthetic sunglasses added.
The results show that the distribution of similarity scores with synthetic sunglasses is shifted to higher similarity compared to the distribution with real sunglasses.
It is possible that the strong specular highlights in the images with real sunglasses provide an ``extra'' lowering of the similarity score.
In any case, these results provide additional confidence that the addition of synthetic sunglasses using FaceLab is an identity-preserving operation.

The resulting dataset, named ``ND-sunglasses", allows us to conduct experiments where the addition of sunglasses is the sole difference between two probe images of a person. To the best of our knowledge, this dataset represents the largest available collection of paired images with and without sunglasses, specifically curated for experiments on one-to-many facial identification.

\section{Experiments}
\label{sec:Experiments}

\label{sec:1-to-Many Accuracy Across Demographics}

The baseline experiment is as follows.  For each person in the dataset, their most recent image is taken as their probe image and their older images are their enrolled images for the gallery.    For each probe image, we compute its rank-one mated similarity score and rank-one non-mated similarity score.  The mated similarity score is the highest similarity between the person’s probe image and one of the same person’s enrolled images.  The non-mated similarity score is the highest similarity between the person’s probe image and an enrolled image of a different person.  If the non-mated similarity score is higher than the mated similarity score, then this probe image results in a rank-one false positive identification (FPI).  To consider results across a set of persons, we look at the distributions of mated and non-mated similarity scores.  The more separated these distributions are, the generally greater the accuracy of one-to-many identification.  To directly focus on the frequency of FPIs, we also consider the distribution of (mated – non-mated) scores across a set of probes.  Each negative value in this distribution is from a probe that generates a false positive identification.  The fraction of such probes gives the FPIR.

\begin{figure}[t]
    \centering
    \begin{subfigure}[b]{1\linewidth}
        \includegraphics[width=\linewidth]{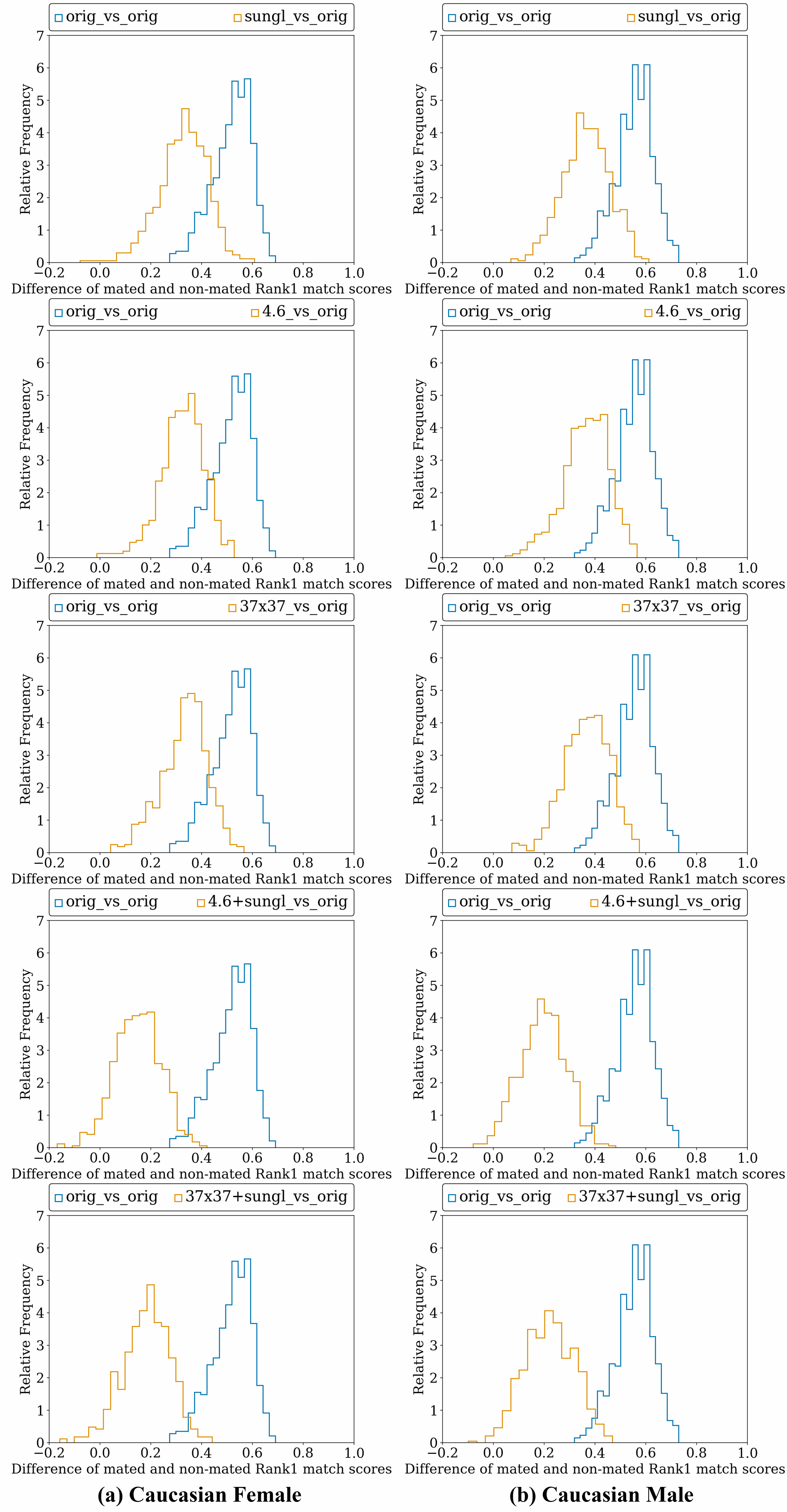}
    \end{subfigure}
    \vspace{-5mm}
   \caption{Scores of differences (mated - non-mated) distributions for original probe images and different conditions of probes (top to bottom): 1) sunglasses, 2) blur of $\sigma = 4.6$, 3) 37x37 resolution, 4) sunglasses + blur, 5) sunglasses + 37x37 resolution. The face matcher is AdaFace.}
\vspace{-5mm}
\label{fig:mated_minus_nonmated}
\end{figure}

\setlength{\tabcolsep}{2mm}

\begin{table*}[]
\tablefontsize{10}
\centering
\begin{tabular}{c|cccc|cccc}
\hline
                                                                                      & \multicolumn{4}{c|}{ArcFace}                                                                                                                                    & \multicolumn{4}{c}{AdaFace}                                                                                                    \\ \cline{2-9} 
                                                                                      & \multicolumn{2}{c|}{CF}                                                              & \multicolumn{2}{c|}{CM}                                         & \multicolumn{2}{c|}{CF}                                              & \multicolumn{2}{c}{CM}                         \\ \cline{2-9} 
\multirow{-3}{*}{Effects added to probe}                                                                    & \multicolumn{1}{c|}{mated}                 & \multicolumn{1}{c|}{non-mated} & \multicolumn{1}{c|}{mated}                 & non-mated & \multicolumn{1}{c|}{mated} & \multicolumn{1}{c|}{non-mated} & \multicolumn{1}{c|}{mated} & non-mated \\ \hline
Sunglasses                                                                   & \multicolumn{1}{c}{\cellcolor[HTML]{9698ED}\textbf{2.6389}} & \multicolumn{1}{c|}{0.3597}             & \multicolumn{1}{c}{\cellcolor[HTML]{9698ED}\textbf{2.8259}} & 0.4723             & \multicolumn{1}{c}{\cellcolor[HTML]{9698ED}\textbf{2.6694}}         & \multicolumn{1}{c|}{0.3003}             & \multicolumn{1}{c}{\cellcolor[HTML]{9698ED}\textbf{2.8283}}         & 0.4357             \\ 
Blur ($\sigma = 4.6$)                                                                     & \multicolumn{1}{c}{\cellcolor[HTML]{9698ED}\textbf{2.7627}} & \multicolumn{1}{c|}{0.1653}             & \multicolumn{1}{c}{\cellcolor[HTML]{9698ED}\textbf{2.7681}} & 0.2128             & \multicolumn{1}{c}{\cellcolor[HTML]{9698ED}\textbf{2.8102}}         & \multicolumn{1}{c|}{0.2099}              & \multicolumn{1}{c}{\cellcolor[HTML]{9698ED}\textbf{2.7823}}         & 0.1506             \\ 
Low resolution (37x37)                                                         & \multicolumn{1}{c}{\cellcolor[HTML]{9698ED}\textbf{2.6253}}  & \multicolumn{1}{c|}{0.2104}             & \multicolumn{1}{c}{\cellcolor[HTML]{9698ED}\textbf{2.7504}} & 0.2569             & \multicolumn{1}{c}{\cellcolor[HTML]{9698ED}\textbf{2.7011}}         & \multicolumn{1}{c|}{0.0243}             & \multicolumn{1}{c}{\cellcolor[HTML]{9698ED}\textbf{3.2738}}         & 0.0686            \\ 
\begin{tabular}[c]{@{}c@{}}Sunglasses + Blur\end{tabular}                                                        & \multicolumn{1}{c}{5.2697}                         & \multicolumn{1}{c|}{0.8537}             & \multicolumn{1}{c}{5.1398}                         & 0.5963             & \multicolumn{1}{c}{5.7830}         & \multicolumn{1}{c|}{0.7960}             & \multicolumn{1}{c}{5.1496}         & 0.6423             \\ 
\begin{tabular}[c]{@{}c@{}}Sunglasses + Low resolution\end{tabular} & \multicolumn{1}{c}{4.4400}                         & \multicolumn{1}{c|}{0.4554}             & \multicolumn{1}{c}{4.5986}                           & 0.5141             & \multicolumn{1}{c}{5.2697}         & \multicolumn{1}{c|}{0.8537}             & \multicolumn{1}{c}{4.5865}         & 0.5505             \\ \hline
\end{tabular}
\caption{The d-prime values are from rank-one similarity distributions. Similar d-prime values are highlighted in \textbf{\textcolor[HTML]{9698ED}{Purple}}. The data demonstrates that adding sunglasses, introducing blur of $\sigma = 4.6$, and reducing resolution to 37x37 in probe images have comparable effects on accuracy. This is notable because, although adding sunglasses to an image might appear to maintain high quality to human observers, it impacts the system's performance similarly to more obvious image degradation.}
\vspace{-4mm}
\label{tab:dprime}
\end{table*}

The baseline experiment is done with the original mugshot-quality images for probes and enrolled gallery.  We then have a series of comparison experiments.  Each comparison experiment involves using a modified version of the probe images, matched against the same original mugshot-quality enrolled gallery.  For example, the core comparison is made by adding sunglasses to each of the probe images, and observing how the accuracy of one-to-many identification changes compared to the baseline experiment.  Making a comparison to the results of the baseline experiment can be done in several ways.  One, we can observe the difference in the distributions of the mated scores, non-mated scores, or (mated – non-mated) scores.  To give a quantitative estimate to the difference in distributions, we compute the d-prime or the Wasserstein distance (also called ``Earth Mover's Distance) \cite{wass-dist} between distributions.  For example, we can compute the d-prime (or Wasserstein) between the baseline (mated – non-mated) distribution and the (mated – non-mated) distribution for the probe images with sunglasses.  A larger d-prime (or Wasserstein) means that sunglasses had a larger effect in degrading accuracy.

The d-prime is commonly used as a quantitative measure of the difference in distributions in biometrics \cite{wu2023beard, wu2022brightness, bhatta-one-to-many}.  However, d-prime assumes that the distributions are approximately Gaussian, and the Wasserstein distance does not.  The Wasserstein distance is not currently commonly used in comparing distributions in biometrics, but could be a better choice especially when the distributions are distinctly not Gaussian.

To consider possible accuracy differences across gender, we compute results separately for male and female.  For fair comparison across gender, because one-to-many accuracy is affected by the number of enrolled persons and images \cite{bhatta-one-to-many}, and there are more males than females in the dataset, we randomly selected 575 Caucasian males to make the number of identities the same for male and female.

We use two face recognition models\footnote{https://github.com/HaiyuWu/SOTA-Face-Recognition-Train-and-Test} pre-trained with the ResNet100 backbone and the Glint360K training set \cite{glint360k}.  One is trained with ArcFace loss \cite{arcface} and one with AdaFace loss \cite{adaface}.  The pattern of accuracy loss due to sunglasses in the probe image is similar between the two matchers.  Due to space limits, some ArcFace results are  in  Supplementary Material.

\subsection{Accuracy Degradation by Sunglasses in Probe}

This section compares one-to-many identification accuracy in two conditions: 
(a) original high-quality probe without sunglasses matched against high-quality gallery without sunglasses, and (b) original high-quality probe image with sunglasses added matched against same gallery as in (a). Unless otherwise specified, (a) is the baseline.

The mated and non-mated distributions for the two demographic groups are shown in the supplementary material. The d-prime values derived from these distributions are presented in Table~\ref{tab:dprime}. In the Adaface results, the d-prime values of mated distributions, 2.6694 for females and 2.8283 for males, suggest a more pronounced effect on males. The d-prime values of non-mated distributions, 0.3003 for females and 0.4357 for males, show that the primary effect is in the mated distributions for probes with sunglasses, both female and male.
Due to the quality of the original images and the size of the dataset, the mated and non-mated 
distributions for probes with sunglasses are still separable, but they have moved closer together.
This shows that the presence of sunglasses in a high-quality probe significantly lowers accuracy.
And, in a real surveillance scenario, images captured from surveillance video often have lower resolution and/or an increased level of blur. Therefore, we explore the effects of combining blur and lower resolution with sunglasses.

\subsection{Effects of Sunglasses + Blur / Resolution}

Since~\cite{bhatta-one-to-many}  comprehensively analyzed the effect of the probe having different degrees of blur and reduced resolution in one-to-many identification experiments, we selected a degree of blur and lower resolution that causes a similar impact on accuracy as  sunglasses. Table~\ref{tab:dprime} shows that when adding $\sigma=4.6$ Gaussian blur to the original probe image or resizing the probe image to 37x37, we obtain a similar d-prime value for the mated distribution as the probe having sunglasses. These were then selected as representative blur and resolution values to investigate combined effects of degraded quality.
The image quality in actual surveillance scenarios can, of course, vary widely.

\setlength{\tabcolsep}{1.9mm}

\begin{table}[h]
\tablefontsize{10}
\centering
\begin{tabular}{c|c|c}
\hline
 Effects added to probe                                                     & Demographic & Distance \\ \hline
\multirow{2}{*}{Sunglasses}                            & \multicolumn{1}{c|}{CF}                  & 0.191                 \\  
                                                                                    & \multicolumn{1}{c|}{CM}                  & 0.190                 \\ \hline
\multirow{2}{*}{Blur ($\sigma = 4.6$)}                 & \multicolumn{1}{c|}{CF}                  & 0.194                 \\  
                                                                                    & \multicolumn{1}{c|}{CM}                  & 0.190                  \\ \hline
\multirow{2}{*}{37x37 resolution}                & \multicolumn{1}{c|}{CF}                  & 0.188                 \\  
                                                                                    & \multicolumn{1}{c|}{CM}                  & 0.187                 \\ \hline
\multirow{2}{*}{Sunglasses + blur}  & \multicolumn{1}{c|}{CF}                  & 0.370                 \\  
                                                                                    & \multicolumn{1}{c|}{CM}                  & 0.360                 \\ \hline
\multirow{2}{*}{Sunglasses + Low resolution } & \multicolumn{1}{c|}{CF}                  & 0.336                 \\  
                                                                                    & \multicolumn{1}{c|}{CM}                  & 0.333                 \\ \hline
\end{tabular}
\vspace{-1mm}
\caption{
Wasserstein distance for shift from baseline caused by effects in the probe images.
Distributions in Figure~\ref{fig:mated_minus_nonmated}. Face matcher is AdaFace.
}
\vspace{-2mm}
\label{tab:wass_prob}
\end{table}

\setlength{\tabcolsep}{1.2mm}

\begin{table}[h]
\centering
\tablefontsize{10}
\begin{tabular}{c|clcl|clcl}
\hline
\multirow{2}{*}{Effects added to probe}                                                                     & \multicolumn{4}{c|}{AdaFace}                               & \multicolumn{4}{c}{ArcFace}                               \\ \cline{2-9} 
                                                                                      & \multicolumn{2}{c|}{CF} & \multicolumn{2}{c|}{CM} & \multicolumn{2}{c|}{CF} & \multicolumn{2}{c}{CM} \\ \hline 
None (Original)                                                                   & \multicolumn{2}{c|}{0}           & \multicolumn{2}{c|}{0}           & \multicolumn{2}{c|}{0}           & \multicolumn{2}{c}{0}           \\ 
Sunglasses                                                                   & \multicolumn{2}{c|}{0.522}           & \multicolumn{2}{c|}{0}           & \multicolumn{2}{c|}{0}           & \multicolumn{2}{c}{0}           \\ 
Blur ($\sigma = 4.6$)                                                                     & \multicolumn{2}{c|}{0.174}           & \multicolumn{2}{c|}{0}           & \multicolumn{2}{c|}{0}    & \multicolumn{2}{c}{0}           \\ 
37x37 resolution                                                         & \multicolumn{2}{c|}{0}    & \multicolumn{2}{c|}{0}           & \multicolumn{2}{c|}{0}    & \multicolumn{2}{c}{0}           \\ 
Sunglasses + blur                                                        & \multicolumn{2}{c|}{4.522}    & \multicolumn{2}{c|}{1.565}    & \multicolumn{2}{c|}{3.652}     & \multicolumn{2}{c}{0.870}    \\ 
\begin{tabular}[c]{@{}c@{}}Sunglasses + Low resolution  \end{tabular} & \multicolumn{2}{c|}{3.304}    & \multicolumn{2}{c|}{0.870}    & \multicolumn{2}{c|}{3.652}    & \multicolumn{2}{c}{0.696}    \\ \hline
\end{tabular}
\vspace{-1mm}
\caption{FPIR for one-to-many identification experiments under various effects in the probe images.
As an example, with sunglasses + blur in the probe images, 4.522\% of Caucasian females probe images had a rank-one false positive identification with AdaFace.}
\vspace{-4mm}
\label{tab:FPIR}
\end{table}

Figure~\ref{fig:images/examples} shows example probe images for six conditions: 1) Original high-quality image, 2) Sunglasses added to original, 3) $\sigma=4.6$ Gaussian blur added to original, 4) Original resized to 37x37, 5) Sunglasses and blur, and 6) Sunglasses and reduced resolution.  Note that adding sunglasses to an original image may subjectively appear to result in a high-quality image, even though accuracy is degraded similar to noticeable blur or low resolution. Table~\ref{tab:dprime} shows that when combining blur and low resolution with sunglasses, the d-prime values are almost doubled between mated distributions and almost tripled between non-mated distributions, compared with the single factor, for both  matchers and both demographic groups. For example, in the AdaFace results for mated females, the d-prime values are 2.6694 when only adding sunglasses to the probes and 5.783 when combining sunglasses and blur in the probes.

\subsection{Analysis of Effect on Model Performance}

To quantitatively analyze the FPIs caused by each condition, we report the Wasserstein Distance of the score difference distribution and the FPIR. 
For a given probe image, a negative score difference indicates an FPI, as the highest similarity value between the probe image and an image from a different person is higher than the top matched image from the same identity. Thus, the distribution of score differences provides a measure of the FPI potential. The distributions of these differences are shown in Figure~\ref{fig:mated_minus_nonmated}. A general conclusion is that each of the probe quality factors -- wearing sunglasses, increased blur, lower resolution -- causes the (mated - non-mated) score distribution to shift to lower values, representing increased chances for FPI. The greatest shifts  occur when probe images have combined blur and sunglasses, or combined lower-resolution and sunglasses.

The Wasserstein Distance~\cite{wass-dist} measures the distance between two distributions without assuming that they are Gaussian. It can be used to compare the size of effect of different probe quality factors on the distribution of (mated - non-mated) scores, 
 with larger values indicating larger degradation in accuracy. 
Table~\ref{tab:wass_prob} shows the Wasserstein Distance between the baseline (original vs. gallery) and each of five instances of degraded probe quality. 
Each of the three individual quality factors (wearing sunglasses, increased blur, decreased resolution) has a similar negative impact on accuracy.
This is to be expected, as the particular blur and resolution values were selected for this.
The sunglasses + blur degradation is close to the sum of the two individual effects, and sunglasses + lower-resolution degradation is slightly below the sum of the two individual effects.
The degree of  accuracy degradation is similar for males and females, although females had lower baseline accuracy.
The overall message from these results is that accuracy degrades rapidly with a combination of different quality factors.

\setlength{\tabcolsep}{1.2mm}
\begin{table}[]
\tablefontsize{10}
\begin{tabular}{c|cccc}
\hline
\multirow{2}{*}{Conditions} & \multicolumn{2}{c|}{CF}                                     & \multicolumn{2}{c}{CM}                 \\ \cline{2-5} 
                                          & \multicolumn{1}{c|}{mated} & \multicolumn{1}{c|}{non-mated} & \multicolumn{1}{c|}{mated} & non-mated \\ \hline
One img. per ID                     & 0.5344                     & \multicolumn{1}{c|}{0.7142}                         & 0.6911                     & 0.8910    \\ \cline{1-1}
All gallery img.                          & \textbf{1.3203}                     & \multicolumn{1}{c|}{\textbf{1.6576}}                         & \textbf{1.6190}                      & \textbf{1.8357}    \\ \hline
\end{tabular}
\vspace{-1mm}
\caption{The d-prime values of rank-one similarity distributions for probe images with sunglasses under different gallery conditions: the first row shows sunglasses randomly added to one image per identity in the gallery, the second row shows sunglasses added to all gallery images. The face matcher is AdaFace.}
\vspace{-4mm}
\label{tab:ada_simi_solu}
\end{table}

We also calculated the FPIR for each probe quality condition, using both  ArcFace and AdaFace. These results in Table~\ref{tab:FPIR} show that both ArcFace and AdaFace have zero FPIR for the original and 37x37 resolution probes. 
AdaFace experiences an FPIR of 0.522\% for sunglasses-added probes and an FPIR of 0.174\% for blurred probes.
But the FPIR increases sharply when sunglasses are combined with either blur or reduced resolution.
For example, sunglasses + blur results in FPIR of 3.652\% for females for ArcFace, and 4.522\% for AdaFace. The absolute level of FPIR varies with the size of a dataset, with a larger number of people enrolled in the gallery causing an increased FPIR \cite{bhatta-one-to-many}. 
So the overall low FPIR in these results does not project what can be experienced for scenarios with larger galleries.
However, the important general point is that multiple degraded quality factors in the probe image causes a dramatic increase in the FPIR.
Also, both matchers consistently show higher FPIR for females than for males. This gender-based difference illustrates another general lesson for one-to-many identification accuracy.
\setlength{\tabcolsep}{5mm}

\begin{table*}[]
\centering
\tablefontsize{10}
\begin{tabular}{c|cccl|cc}
\hline
                                                                                                                                    & \multicolumn{4}{c|}{\begin{tabular}[c]{@{}c@{}}Shift in mated - non-mated \\ compared to baseline\end{tabular}} & \multicolumn{2}{c}{\% of lost accuracy recovered}                                                       \\ \cline{2-7} 
\multirow{-2}{*}{Experiment Conditions}                                                                                             & \multicolumn{1}{c}{}                & \multicolumn{1}{c}{AdaFace}              & \multicolumn{2}{c|}{ArcFace} & \multicolumn{1}{c}{AdaFace}                                  & ArcFace                                  \\ \hline
                                                                                                                                    & \multicolumn{1}{c}{CF}              & \multicolumn{1}{c}{0.191}                & \multicolumn{2}{c|}{0.197}   & \multicolumn{1}{c}{-}                                        & -                                        \\ 
\multirow{-2}{*}{sunglasses probe vs. original gallery}                                                                             & \multicolumn{1}{c}{CM}              & \multicolumn{1}{c}{0.190}                & \multicolumn{2}{c|}{0.194}   & \multicolumn{1}{c}{-}                                        & -                                        \\ \hline
                                                                                                                                    & \multicolumn{1}{c}{CF}              & \multicolumn{1}{c}{0.171}                & \multicolumn{2}{c|}{0.176}   & \multicolumn{1}{c}{\cellcolor[HTML]{34CDF9}\textbf{10.47\%}} & \cellcolor[HTML]{34CDF9}\textbf{10.66\%} \\  
\multirow{-2}{*}{\begin{tabular}[c]{@{}c@{}}sunglasses probe vs. add sunglasses to \\ one image per ID in the gallery\end{tabular}} & \multicolumn{1}{c}{CM}              & \multicolumn{1}{c}{0.164}                & \multicolumn{2}{c|}{0.163}   & \multicolumn{1}{c}{\cellcolor[HTML]{CBCEFB}\textbf{13.68\%}} & \cellcolor[HTML]{CBCEFB}\textbf{15.98\%} \\ \hline
                                                                                                                                    & \multicolumn{1}{c}{CF}              & \multicolumn{1}{c}{0.136}                & \multicolumn{2}{c|}{0.141}   & \multicolumn{1}{c}{\cellcolor[HTML]{32CB00}\textbf{28.80\%}}  & \cellcolor[HTML]{32CB00}\textbf{28.43\%} \\  
\multirow{-2}{*}{\begin{tabular}[c]{@{}c@{}}sunglasses probe vs. add sunglasses to \\all gallery images\end{tabular}}                                                         & \multicolumn{1}{c}{CM}              & \multicolumn{1}{c}{0.122}                & \multicolumn{2}{c|}{0.121}   & \multicolumn{1}{c}{\cellcolor[HTML]{F8A102}\textbf{35.79\%}} & \cellcolor[HTML]{F8A102}\textbf{37.63\%} \\ \hline
\end{tabular}
\vspace{-1mm}
\caption{Wasserstein Distance of the differences in (mated - non-mated) distributions. Adding sunglasses to one image per ID in the gallery recovered 10.47\% with AdaFace and 10.66\% with ArcFace \textbf{\textcolor[HTML]{34CDF9}{Blue}} of the accuracy lost for females and 13.68\% with AdaFace and 15.98\% with ArcFace \textbf{\textcolor[HTML]{CBCEFB}{Purple}} for males. Adding sunglasses to all gallery images recovered 28.80\% with AdaFace and 28.43\% with ArcFace \textbf{\textcolor[HTML]{32CB00}{Green}} of the accuracy lost for females and 35.79\% with AdaFace and 37.63\% with ArcFace \textbf{\textcolor[HTML]{F8A102}{Orange}} for males.}
\vspace{-4mm}
\label{tab:wass_solu}
\end{table*}

\subsection{Strategy for Recovering Accuracy Loss}

Previous sections show that a probe image with sunglasses matched against gallery images without sunglasses results in a substantial decrease in accuracy.
Primarily, the occlusion of facial features by sunglasses in the probe image obscures vital identifying information. 
But also, comparing images with and without sunglasses presents a mismatch in facial features, which is potentially a different factor affecting accuracy. 
One could imagine an approach to remove sunglasses from the probe image, or an approach to add sunglasses to the gallery images.
Removing sunglasses from the probe could perhaps require less computation, as it involves the one image, but it requires hallucinating face feature content that is unseen.
Adding sunglasses to the gallery images, on the other hand, involves adding a plausible instance of a specific style of occlusion to each image.

In this section, we investigate whether adding sunglasses to the gallery images can mitigate the accuracy degradation that would otherwise occur when the probe has sunglasses. 

\begin{figure}[t]
    \centering
    \begin{subfigure}[b]{1\linewidth}
        \includegraphics[width=\linewidth]{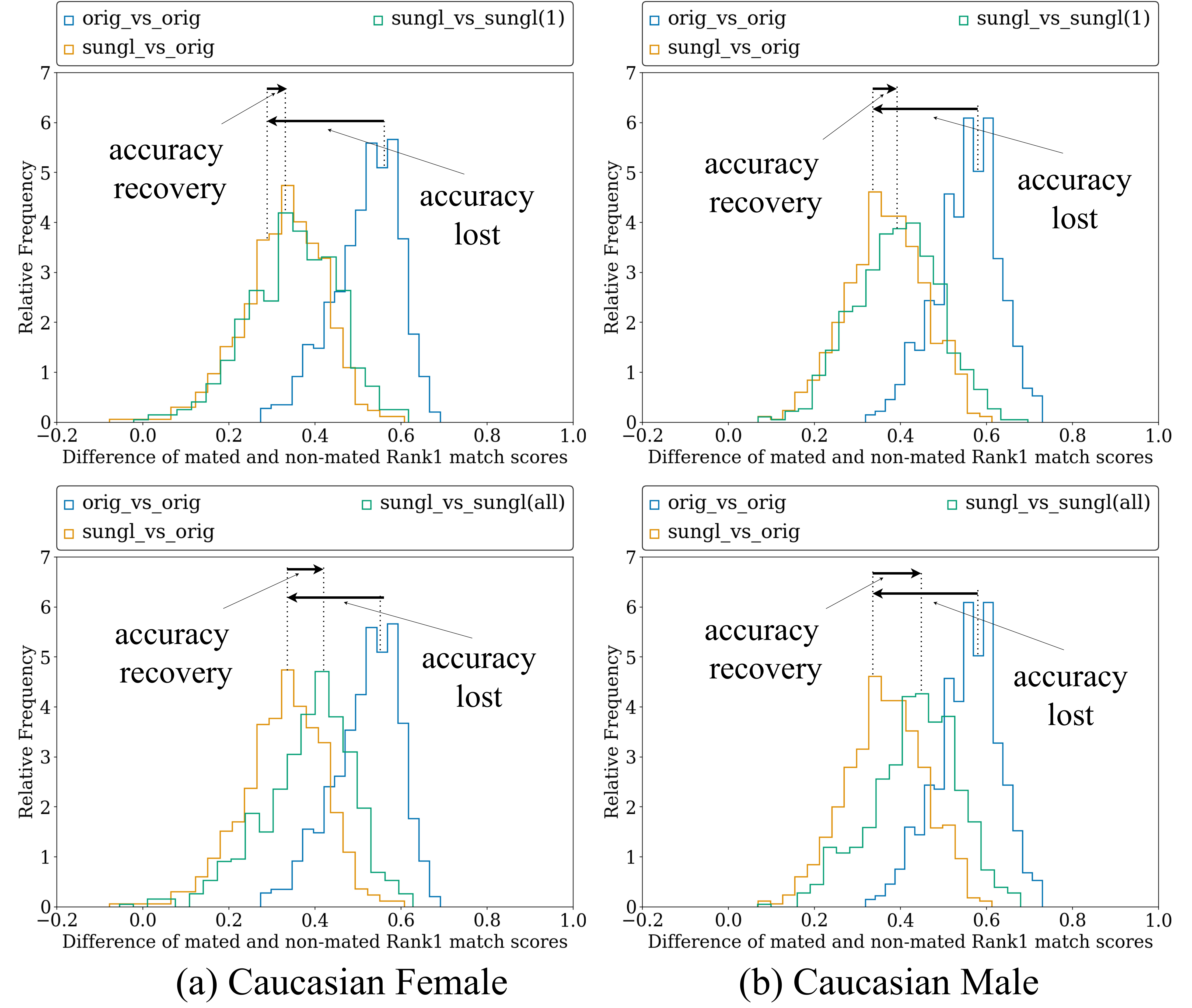}
    \end{subfigure}
    \vspace{-5mm}
   \caption{Similarity difference (mated - non-mated) distributions. Comparing with the \textcolor[HTML]{0c80d5}{baseline}, adding sunglasses to probe images causes a large \textit{accuracy loss}. To counter this effect, we add sunglasses to one (all) image(s) per identity in the gallery to achieve good \textit{accuracy recovery}, as shown in Top (Bottom) row. The face matcher is AdaFace.}
\vspace{-4mm}
\label{fig:diff_solu}
\end{figure}

Table~\ref{tab:ada_simi_solu} shows the d-prime values of rank-one similarity distributions for probe images with sunglasses under different gallery conditions. When sunglasses are randomly added to a single image per identity in the gallery, with an average of 11.15 images per identity, the results show d-prime values of 0.5344 (mated) and 0.7142 (non-mated) for females, and 0.6911 (mated) and 0.891 (non-mated) for males. The d-prime values become more pronounced when sunglasses are added to all gallery images:
1.3203 (mated) and 1.6576 (non-mated) for females, and 1.619 (mated) and 1.8357 (non-mated) for males. Notably, in both scenarios, the improvement in accuracy is more substantial for males than for females. The results demonstrate that under both gallery conditions, accuracy improves. However, adding sunglasses to all gallery images recovers more accuracy. The value of adding sunglasses to the gallery images to ameliorate the accuracy degradation is more obvious in the (mated - non-mated) distributions shown in Figure~\ref{fig:diff_solu}. 
 
Table~\ref{tab:wass_solu} presents the Wasserstein Distances of the difference distributions. 
When we randomly select one image per identity in the gallery to add sunglasses, 1150 images in total, it recovers 10.47\% with AdaFace and 10.66\% with ArcFace of the lost accuracy for females and 13.68\% with AdaFace and 15.98\% with ArcFace for males when the person in the probe image is wearing sunglasses. Furthermore, when sunglasses are added to all gallery images, we observe a more substantial recovery of 28.8\% with AdaFace and 28.43\% with ArcFace of the lost accuracy for females and 35.79\% with AdaFace and 37.63\% with ArcFace for males. It's worth noting that both the d-prime and Wasserstein Distance metrics showed similar general patterns in our results. However, visual inspection suggests the distributions are not strictly Gaussian. Given this, the Wasserstein Distance may be a more appropriate metric for this type of study, even though in this case the overall pattern of results was not substantially different between the two measures.

It is important to note that 100\% recovery of lost accuracy is not feasible due to the occlusion of  facial features by sunglasses and because the type of sunglasses in the gallery is not matched to those in the probe, so even for the same identity,  probe and gallery images may contain different sunglasses. Nevertheless, our results show the potential of this approach to recover part of the lost  accuracy without re-training the models, thereby enhancing the robustness of one-to-many facial identification systems in scenarios involving occlusions.
In a real-world scenario, it may be more advantageous to consistently add sunglasses similar to those encountered in probe images to the gallery images.

\subsection{Effect of Sunglasses Training Samples in Accuracy}
Accuracy lost by adding sunglasses to face can possibly be explained by the under-representation of wearing-sunglasses images in the training set. To explore this, we start with a widely recognized training set, WebFace4M~\cite{webface260m}, which has 4M images from 200K identities.

WebFace4M images do not come with meta-data for sunglasses.
And we did not find a suitable sunglasses detector already available. 
Therefore, we developed a sunglasses detector to estimate the number of wearing-sunglasses samples in WebFace4M. Specifically, we first use a facial attribute classifier~\cite{logicnet} to get the wearing-eyeglasses images in WebFace4M. Part of these images are then manually separated to three classes: no-eyeglasses, eyeglasses, and sunglasses, for the next training step. This results in 8,500 dark sunglasses images, 8,200 eyeglasses images, and 8,200 no-eyeglasses images. This dataset is subsequently partitioned into training, validation, and test allocations of 70\%, 15\%, and 15\%, respectively. We then fine-tuned a ResNet101~\cite{resnet}, pre-trained on ImageNet, using this dataset. This sunglasses detector achieves 93.3\% accuracy on the test set. After running this detector on WebFace4M, 115,818 (3.8\%) images are predicted as having dark sunglasses. 
The strong under-representation of sunglasses in the training set motivates increasing the representation of wearing-sunglasses images in the training set as a possible way to improve accuracy.

\setlength{\tabcolsep}{1.3mm}

\begin{table}[]
\tablefontsize{10}
\begin{tabular}{c|cccc}
\hline
\multirow{2}{*}{Sunglasses Types} & \multicolumn{4}{c}{sunglasses images (\%)}                                                           \\ \cline{2-5} 
                                  & \multicolumn{1}{c|}{0\%}     & \multicolumn{1}{c|}{10\%}    & \multicolumn{1}{c|}{20\%}    & 23\%    \\ \hline
real sunglasses                   & \multicolumn{1}{c|}{95.31\%} & \multicolumn{1}{c|}{95.18\%} & \multicolumn{1}{c|}{95.05\%} & 95.22\% \\
synthetic sunglasses              & \multicolumn{1}{c|}{95.31\%} & \multicolumn{1}{c|}{95.24\%} & \multicolumn{1}{c|}{95.11\%} & 95.17\% \\ \hline
\end{tabular}
\vspace{-1mm}
\caption{A general face recognition test accuracy (\%) on real and synthetic sunglasses datasets across models trained with varying proportions (0\%, 10\%, 20\%, and 23\%) of sunglasses-augmented images.}
\vspace{-5mm}
\label{tab:train}
\end{table}

\begin{figure}[t]
    \centering
    \begin{subfigure}[b]{1\linewidth}
        \includegraphics[width=\linewidth]{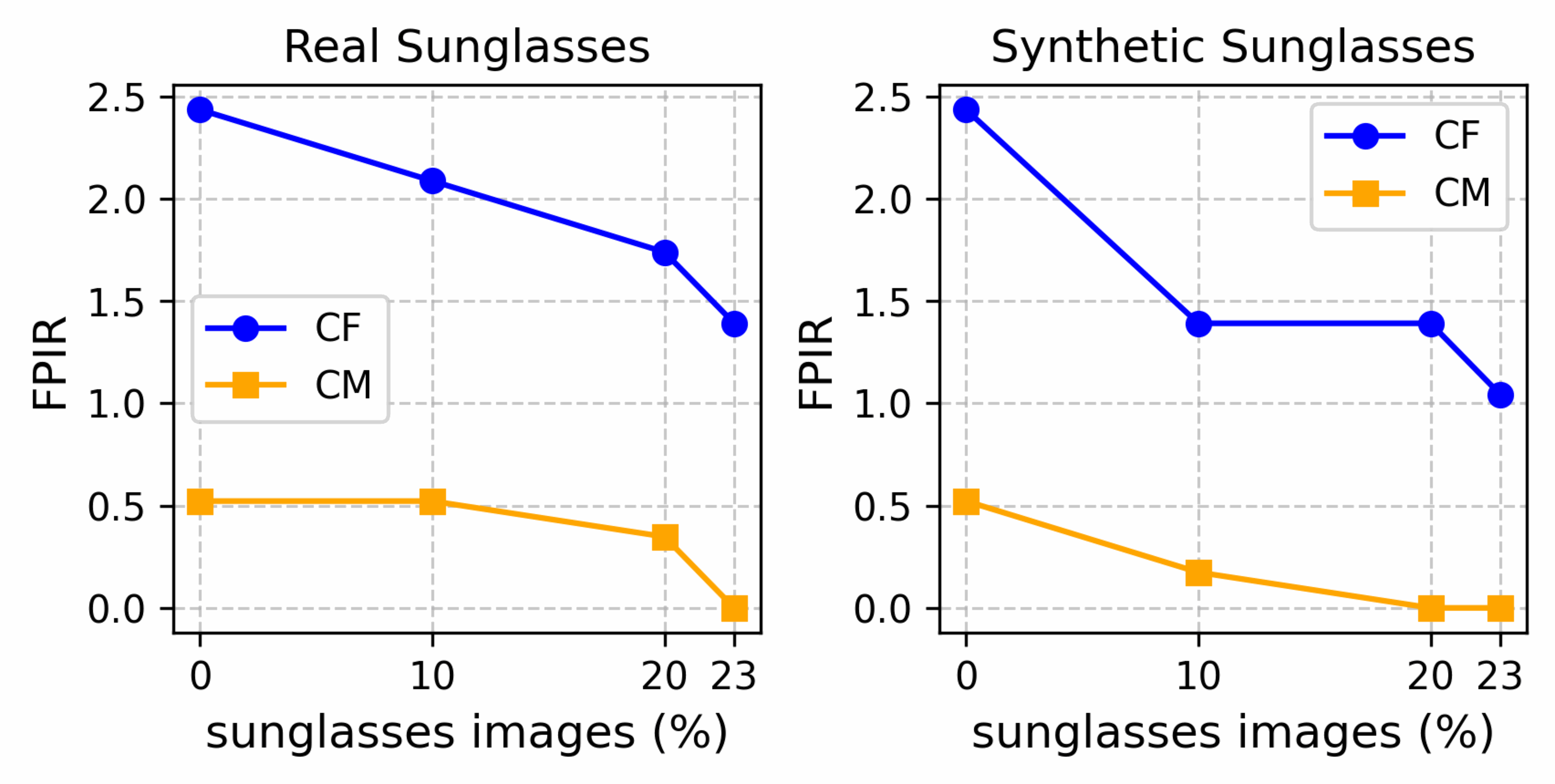}
    \end{subfigure}
    \vspace{-5mm}
   \caption{This illustrates FPIR changes as the proportion of sunglasses-augmented images in the training set from 0\% to 23\%, for both real (left) and synthetic (right) sunglasses. Performance is evaluated in one-to-many identification experiments where probe images are wearing sunglasses.}

\label{fig:train}
\end{figure}

However, adding sunglasses images to the dataset is non-trivial, so we investigate how the fraction of sunglasses images impacts the accuracy. First, we downscale the dataset to 500K by randomly selecting the no-sunglasses images from 38,802 identities that are detected as having sunglasses images. In this scale, the detected wearing-sunglasses images occupy roughly 23\% of the dataset. We then train the models with varying proportions of wearing-sunglasses images by replacing the no-sunglasses images with wearing-sunglasses images. These models are tested on ND-Sunglasses to obtain the FPIR trend.
The left plot in Figure~\ref{fig:train} indicates a clear decreasing FPIR when increasing the fraction of sunglasses images. From 0\% to 23\%, the FPIR has a 42.87\% decrease for CF and all CM probe images are correctly identified. 

\begin{figure}[t]
    \centering
    \begin{subfigure}[b]{1\linewidth}
        \includegraphics[width=\linewidth]{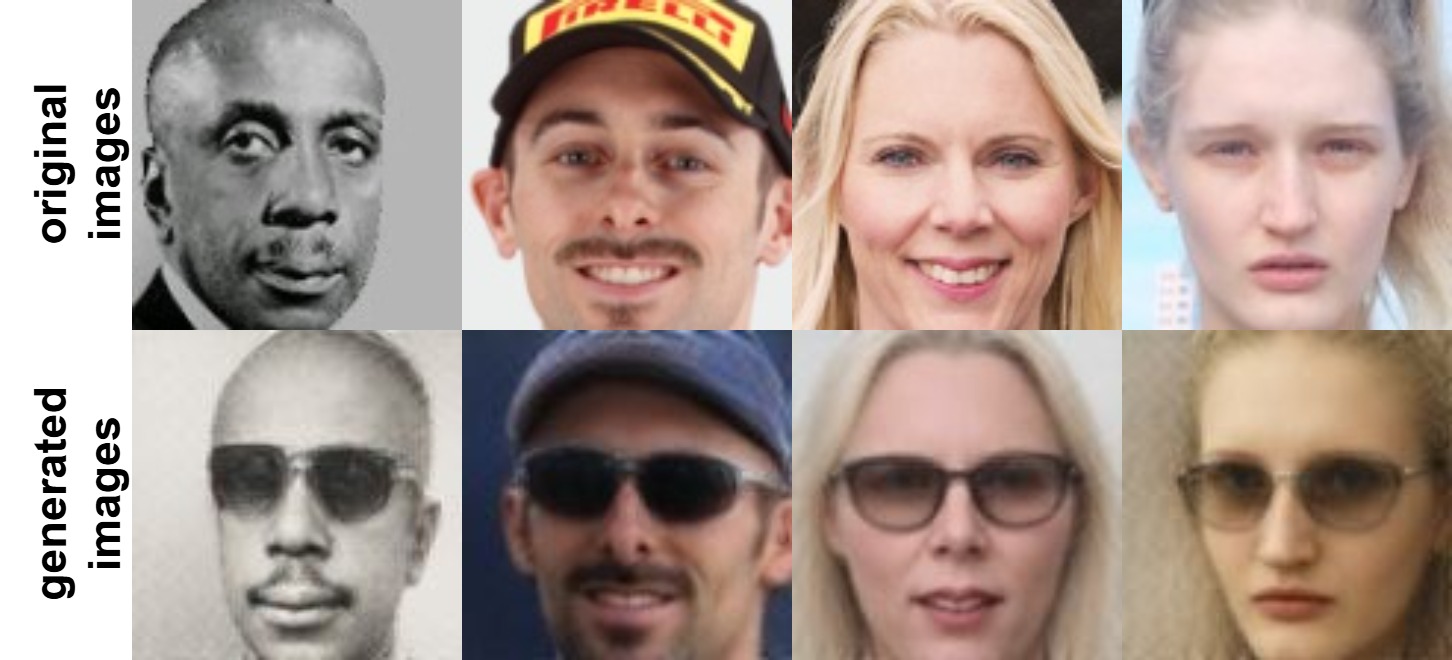}
    \end{subfigure}
    \vspace{-5mm}
   \caption{Examples of original images (top row) and generated images with sunglasses (bottom row) using the Vec2Face generative model.}
\vspace{-5mm}
\label{fig:train_examples}
\end{figure}

To efficiently add new wearing-sunglasses images to the dataset, we choose to use a face generative model, Vec2Face~\cite{vec2face}. It can effectively generate images of the same identity with sunglasses, as shown in Figure~\ref{fig:train_examples}. We then conduct the same experiments to evaluate if the synthesized wearing-sunglasses images can achieve the similar effect to the real ones. Note that we uniformly generate images for all 38,082 identities. The right plot in Figure~\ref{fig:train} indicates that using synthesized images can achieve better performance than real ones. The FPIR has a 57.17\% decrease for CF and all CM probe images are correctly identified. This showcases the possible accuracy improvement in the full-size WebFace4M by adding synthesized wearing-sunglasses images in the future work. It is worth noting that all the trained models have close average accuracy on five standard test sets~\cite{lfw, cplfw, calfw, agedb-30, cfpfp}, as shown in Table~\ref{tab:train}.

\section{Conclusions and Future Work}
\label{sec:Conclusions and Future Work}

This study systematically investigates the impact of sunglasses in the probe images on one-to-many identification accuracy, addressing a gap in existing literature. We assembled a dataset, ND-Sunglasses, to support the investigation.
Our findings reveal that sunglasses in probe images substantially reduces the accuracy in one-to-many identification, comparable to applying Gaussian blur with $\sigma = 4.6$ or reducing image resolution to 37x37 pixels, despite these latter factors causing more perceptible visual changes. 

Our research also demonstrates that generating sunglasses-wearing versions of gallery images can partially mitigate the accuracy loss caused by sunglasses in probe images. Specifically, adding sunglasses to all gallery images recovered 35.79\% of lost accuracy for females and 37.63\% for males with ArcFace. 

We show that the accuracy degradation for probe images with sunglasses can also be addressed by fixing the under-representation of wearing-sunglasses images in the training data. We show that using a face generative model to uniformly add wearing-sunglasses images to the training set can reduce the FPIR. Future work is to explore versions of this strategy on a full-size training set.

\section{Ethical Impact Statement}

Documented instances of wrongful arrest have brought critical attention to one-to-many facial identification.
This research improves understanding how quality factors in a probe image degrade accuracy, and should lead to more realistic appraisal of the accuracy of one-to-many identification,  hopefully contributing to reducing instances of wrongful arrest.

\newpage
{\small
\bibliographystyle{ieee}
\bibliography{egbib}

\begin{thebibliography}{10}\itemsep=-1pt

\bibitem{Albiero_IJCB_2022}
V.~Albiero, K.~W. Bowyer, and M.~C. King.
\newblock Face regions impact recognition accuracy differently across demographics.
\newblock In {\em IJCB}, 2022.

\bibitem{img2pose}
V.~Albiero, X.~Chen, X.~Yin, G.~Pang, and T.~Hassner.
\newblock img2pose: Face alignment and detection via 6dof, face pose estimation.
\newblock {\em CVPR}, pages 7613--7623, 2020.

\bibitem{glint360k}
X.~An, X.~Zhu, Y.~Gao, Y.~Xiao, Y.~Zhao, Z.~Feng, L.~Wu, B.~Qin, M.~Zhang, D.~Zhang, and Y.~Fu.
\newblock Partial {FC:} training 10 million identities on a single machine.
\newblock In {\em ICCVW}, pages 1445--1449, 2021.

\bibitem{Bhatta_WACVW_2023}
A.~Bhatta, V.~Albiero, K.~W. Bowyer, and M.~C. King.
\newblock The gender gap in face recognition accuracy is a hairy problem.
\newblock In {\em WACVW}, 2023.

\bibitem{tinyface}
Z.~Cheng, X.~Zhu, and S.~Gong.
\newblock Low-resolution face recognition.
\newblock In {\em ACCV}, pages 605--621, 2019.

\bibitem{arcface}
J.~Deng, J.~Guo, N.~Xue, and S.~Zafeiriou.
\newblock Arcface: Additive angular margin loss for deep face recognition.
\newblock In {\em CVPR}, pages 4690--4699, 2019.

\bibitem{sunglass-work}
M.~E. Erak$\iota$n, U.~Demir, and H.~K. Ekenel.
\newblock On recognizing occluded faces in the wild.
\newblock In {\em 2021 International Conference of the Biometrics Special Interest Group (BIOSIG)}, pages 1--5. IEEE, 2021.

\bibitem{wrong-arrest-Murphy}
B.~Fung.
\newblock Lawsuit: Facial recognition software leads to wrongful arrest of texas man; he was in sacramento at time of robbery.
\newblock {\em {CBS News}}, January 23, 2024.

\bibitem{Grother_NIST_2019b}
P.~Grother, M.~Ngan, and K.~Hanaoka.
\newblock Face recognition vendor test ({FRVT}) part 2: Identification.
\newblock {\em NISTIR}, 8271, 2019.

\bibitem{Grother_NIST_2019}
P.~Grother, M.~Ngan, and K.~Hanaoka.
\newblock Ongoing face recognition vendor test ({FRVT}) part 3: Demographic effects.
\newblock {\em NISTIR}, 8280, 2019.

\bibitem{meglass}
J.~Guo, X.~Zhu, Z.~Lei, and S.~Z. Li.
\newblock Face synthesis for eyeglass-robust face recognition.
\newblock In {\em CCBR}, pages 275--284, 2018.

\bibitem{resnet}
K.~He, X.~Zhang, S.~Ren, and J.~Sun.
\newblock Deep residual learning for image recognition.
\newblock In {\em CVPR}, pages 770--778, 2016.

\bibitem{attgan}
Z.~He, W.~Zuo, M.~Kan, S.~Shan, and X.~Chen.
\newblock Attgan: Facial attribute editing by only changing what you want.
\newblock {\em IEEE transactions on image processing}, 28(11):5464--5478, 2019.

\bibitem{wass-dist}
L.~Hou, C.-P. Yu, and D.~Samaras.
\newblock Squared earth mover's distance-based loss for training deep neural networks.
\newblock {\em arXiv preprint arXiv:1611.05916}, 2016.

\bibitem{lfw}
G.~B. Huang, M.~Mattar, T.~Berg, and E.~Learned-Miller.
\newblock Labeled faces in the wild: A database forstudying face recognition in unconstrained environments.
\newblock In {\em Workshop on faces in'Real-Life'Images: detection, alignment, and recognition}, 2008.

\bibitem{wrong-arrest-WilliamsOliverParks}
K.~Johnson.
\newblock How wrongful arrests based on {AI} derailed 3 men's lives.
\newblock {\em {Wired News}}, March 7, 2022.

\bibitem{adaface}
M.~Kim, A.~K. Jain, and X.~Liu.
\newblock Adaface: Quality adaptive margin for face recognition.
\newblock In {\em CVPR}, pages 18729--18738, 2022.

\bibitem{ijba}
B.~F. Klare, B.~Klein, E.~Taborsky, A.~Blanton, J.~Cheney, K.~Allen, P.~Grother, A.~Mah, M.~J. Burge, and A.~K. Jain.
\newblock Pushing the frontiers of unconstrained face detection and recognition: {IARPA} janus benchmark {A}.
\newblock In {\em CVPR}, 2015.

\bibitem{Krishnapriya_TTS_2020}
K.~Krishnapriya, V.~Albiero, K.~Vangara, M.~King, and K.~Bowyer.
\newblock Issues related to face recognition accuracy varying based on race and skin tone.
\newblock 2020.

\bibitem{stgan}
M.~Liu, Y.~Ding, M.~Xia, X.~Liu, E.~Ding, W.~Zuo, and S.~Wen.
\newblock Stgan: A unified selective transfer network for arbitrary image attribute editing.
\newblock In {\em Proceedings of the IEEE/CVF conference on computer vision and pattern recognition}, pages 3673--3682, 2019.

\bibitem{Martinez_AR_1998}
A.~Martinez and R.~Benavente.
\newblock The {AR} face database.
\newblock {\em CVC TechRep \#24}, 1998.

\bibitem{ijbc}
B.~Maze, J.~C. Adams, J.~A. Duncan, N.~D. Kalka, T.~Miller, C.~Otto, A.~K. Jain, W.~T. Niggel, J.~Anderson, J.~Cheney, and P.~Grother.
\newblock {IARPA} janus benchmark - {C:} face dataset and protocol.
\newblock In {\em International Conference on Biometrics}, pages 158--165, 2018.

\bibitem{agedb-30}
S.~Moschoglou, A.~Papaioannou, C.~Sagonas, J.~Deng, I.~Kotsia, and S.~Zafeiriou.
\newblock Agedb: The first manually collected, in-the-wild age database.
\newblock In {\em CVPRW}, pages 1997--2005, 2017.

\bibitem{bhatta-one-to-many}
G.~Pangelinan, A.~Bhatta, H.~Wu, M.~C. King, and K.~W. Bowyer.
\newblock Analyzing the impact of demographic and operational variables on 1-to-many face id search.
\newblock {\em IEEE Transactions on Technology and Society}, 2024.

\bibitem{notre-dame}
P.~J. Phillips, P.~J. Flynn, W.~T. Scruggs, K.~Bowyer, J.~Chang, K.~Hoffman, J.~Marques, J.~Min, and W.~J. Worek.
\newblock Overview of the face recognition grand challenge.
\newblock {\em CVPR}, 1:947--954 vol. 1, 2005.

\bibitem{morph}
K.~Ricanek and T.~Tesafaye.
\newblock Morph: A longitudinal image database of normal adult age-progression.
\newblock In {\em IEEE Face \& Gesture Recognition}, pages 341--345, 2006.

\bibitem{cfpfp}
S.~Sengupta, J.~Chen, C.~D. Castillo, V.~M. Patel, R.~Chellappa, and D.~W. Jacobs.
\newblock Frontal to profile face verification in the wild.
\newblock In {\em WACV}, pages 1--9, 2016.

\bibitem{eyeglasses-add}
W.~Shen and R.~Liu.
\newblock Learning residual images for face attribute manipulation.
\newblock In {\em CVPR}, pages 4030--4038, 2017.

\bibitem{wrong-arrest-Reid}
S.~Thanawala.
\newblock Facial recognition technology jailed a man for days. his lawsuit joins others from black plaintiffs.
\newblock {\em {Associated Press}}, September 25, 2023.

\bibitem{ND_MFAD}
M.~C.~K. Vitor~Albiero, Kevin W.~Bowyer.
\newblock Notre dame male/female accuracy dataset.

\bibitem{wang2016recognizing}
T.~Y. Wang and A.~Kumar.
\newblock Recognizing human faces under disguise and makeup.
\newblock In {\em 2016 IEEE International Conference on Identity, Security and Behavior Analysis (ISBA)}, pages 1--7. IEEE, 2016.

\bibitem{ijbb}
C.~Whitelam, E.~Taborsky, A.~Blanton, B.~Maze, J.~C. Adams, T.~Miller, N.~D. Kalka, A.~K. Jain, J.~A. Duncan, K.~Allen, J.~Cheney, and P.~Grother.
\newblock {IARPA} janus benchmark-b face dataset.
\newblock In {\em CVPRW}, pages 592--600, 2017.

\bibitem{wu2022brightness}
H.~Wu, V.~Albiero, K.~Krishnapriya, M.~C. King, and K.~W. Bowyer.
\newblock Face recognition accuracy across demographics: Shining a light into the problem.
\newblock In {\em CVPR}, pages 1041--1050, 2023.

\bibitem{wu2023beard}
H.~Wu, G.~Bezold, A.~Bhatta, and K.~W. Bowyer.
\newblock Logical consistency and greater descriptive power for facial hair attribute learning.
\newblock In {\em CVPR}, pages 8588--8597, 2023.

\bibitem{vec2face}
H.~Wu, J.~Singh, S.~Tian, L.~Zheng, and K.~W. Bowyer.
\newblock Vec2face: Scaling face dataset generation with loosely constrained vectors.
\newblock {\em arXiv preprint arXiv:2409.02979}, 2024.

\bibitem{hadrian-eclipse}
H.~Wu, S.~Tian, A.~Bhatta, J.~Gutierrez, G.~Bezold, G.~Argueta, K.~Ricanek, M.~C. King, and K.~W. Bowyer.
\newblock What is a goldilocks face verification test set?
\newblock {\em ArXiv}, abs/2405.15965, 2024.

\bibitem{logicnet}
H.~Wu, S.~Tian, H.~Li, and K.~W. Bowyer.
\newblock Logicnet: A logical consistency embedded face attribute learning network.
\newblock {\em WACV}, 2025.

\bibitem{wrong-arrest-Woodruff}
I.~Yip.
\newblock Detroit police chief says ‘poor investigative work’ led to arrest of black mom who claims facial recognition technology played a role.
\newblock {\em {CNN}}, August 10, 2023.

\bibitem{cplfw}
T.~Zheng and W.~Deng.
\newblock Cross-pose {LFW}: A database for studying cross-pose face recognition in unconstrained environments.
\newblock {\em Beijing University of Posts and Telecommunications, Tech. Rep}, 5(7), 2018.

\bibitem{calfw}
T.~Zheng, W.~Deng, and J.~Hu.
\newblock Cross-age {LFW}: A database for studying cross-age face recognition in unconstrained environments.
\newblock {\em arXiv preprint arXiv:1708.08197}, 2017.

\bibitem{webface260m}
Z.~Zhu, G.~Huang, J.~Deng, Y.~Ye, J.~Huang, X.~Chen, J.~Zhu, T.~Yang, D.~Du, J.~Lu, and J.~Zhou.
\newblock Webface260m: {A} benchmark for million-scale deep face recognition.
\newblock {\em PAMI}, 45(2):2627--2644, 2023.

\end{thebibliography}
}

\clearpage

\twocolumn[{\centering
\Large\textbf{Supplementary Material\\[3mm]}}]

\begin{figure}[t]
    \centering
    \begin{subfigure}[b]{1\linewidth}
        \includegraphics[width=\linewidth]{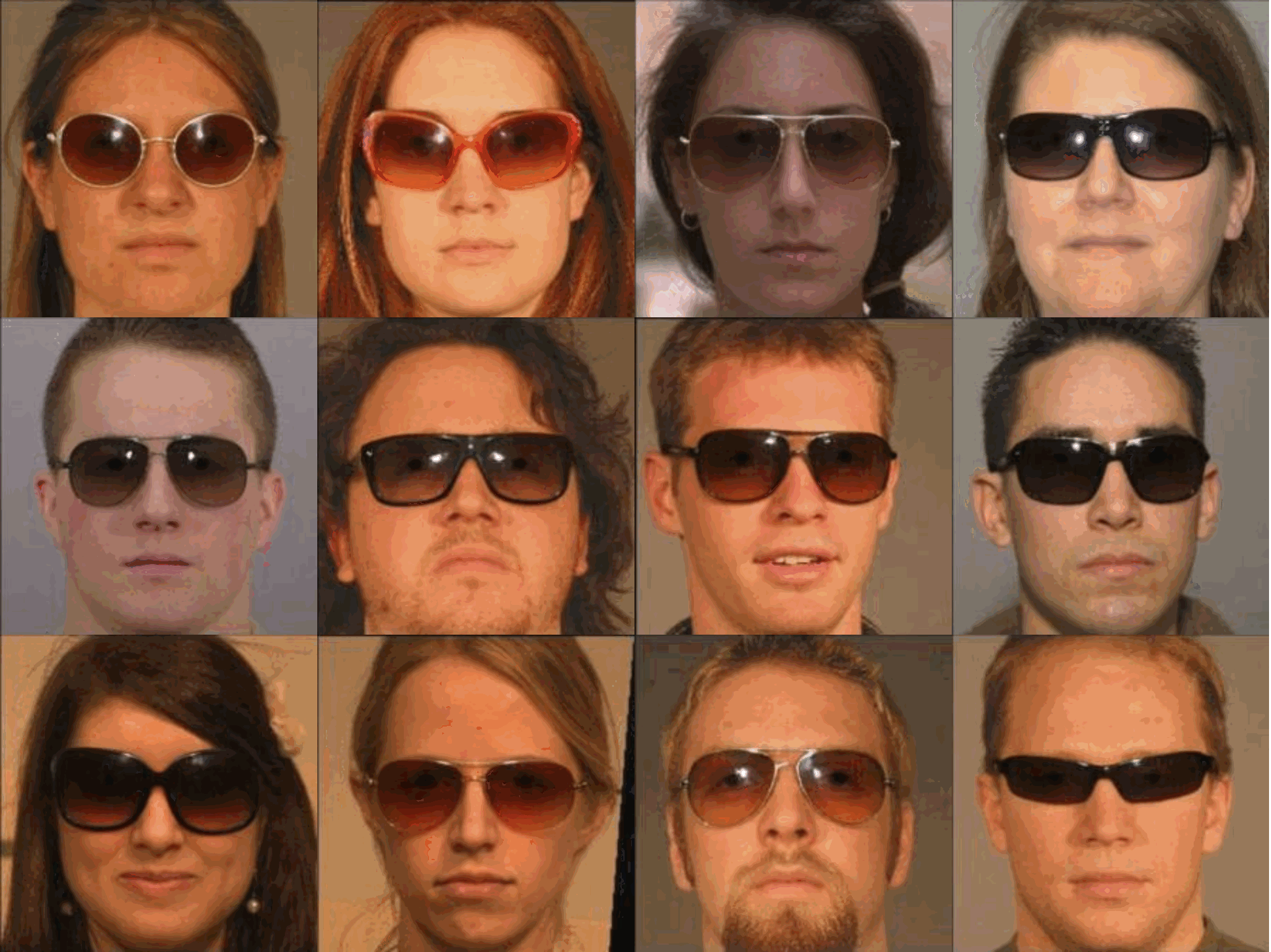}
    \end{subfigure}
    \vspace{-3mm}
   \caption{Examples of different types of synthetic sunglasses added by FaceLab app.}
\vspace{2mm}
\label{fig:diff_types}
\end{figure}

\begin{figure}[t]
    \centering
        \includegraphics[width=\linewidth]{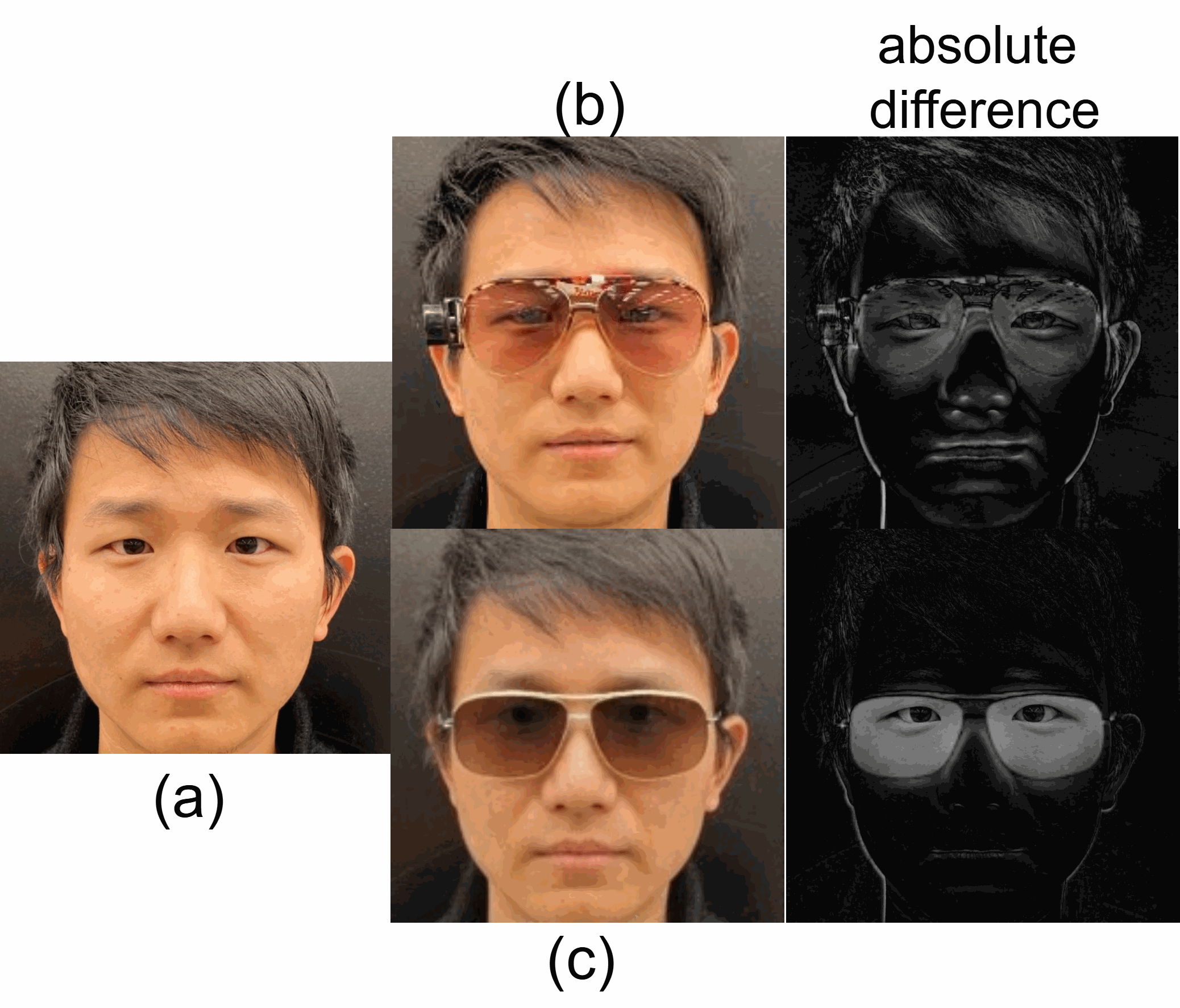}
   \caption{This study illustrates a small experiment comprising three conditions: a) the original image, b) the image with physical sunglasses applied, and c) the image with sunglasses using the FaceLab app. Absolute differences between these images were calculated, along with cosine similarities between the original and each modified version. Results revealed cosine similarities of 0.773 for physical sunglasses and 0.819 for FaceLab-applied sunglasses. The higher similarity of FaceLab-augmented images to the original supports the validity of using the FaceLab application for sunglasses augmentation in our research methodology.}
   \vspace{-4mm}
\label{fig:images/sup_small_experiment}
\end{figure}

\begin{figure}[t]
    \centering
    \begin{subfigure}[b]{1\linewidth}
        \includegraphics[width=\linewidth]{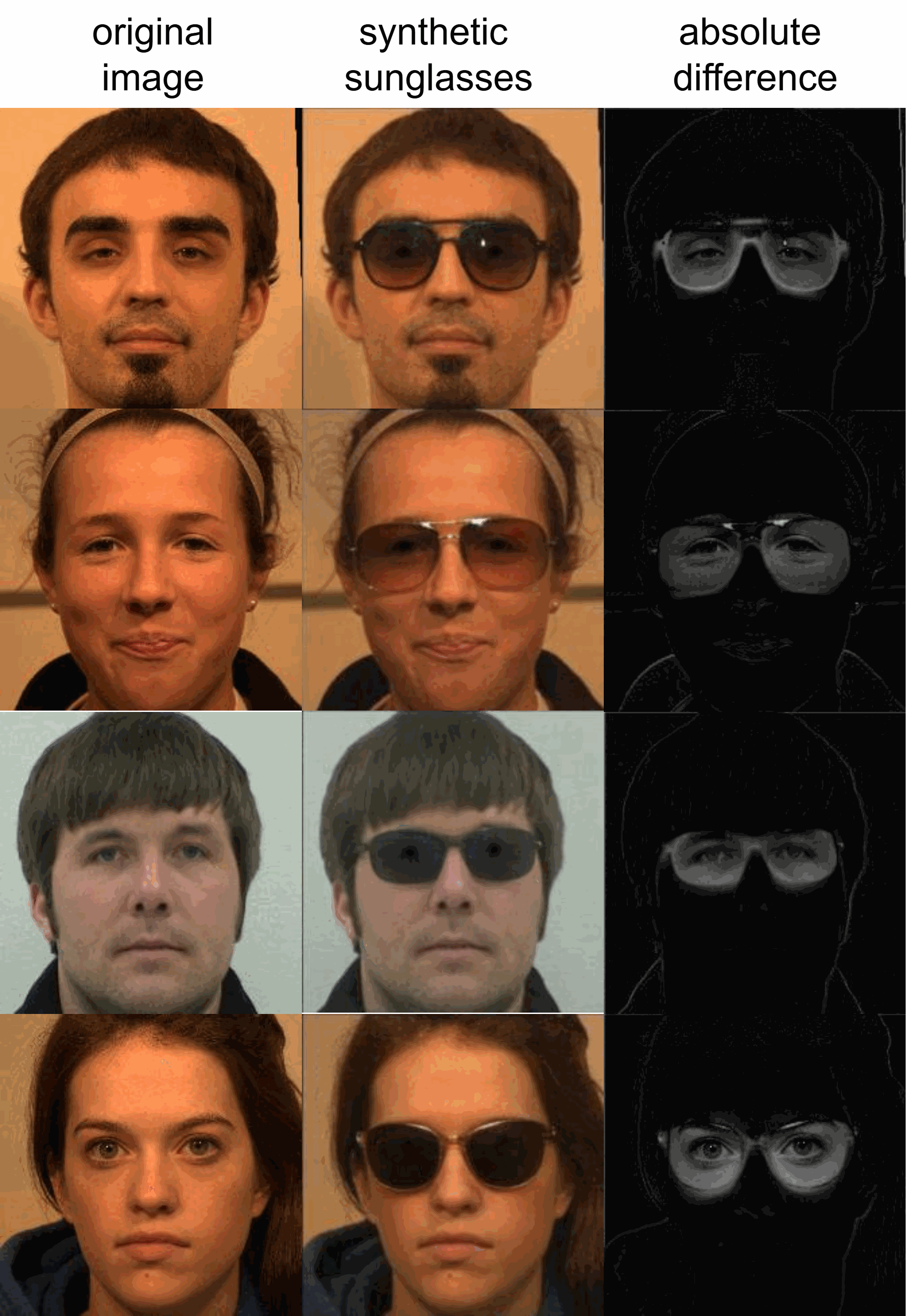}
    \end{subfigure}
    \vspace{-3mm}
   \caption{The figure comprises three columns: the original image, the FaceLab app-modified image with added sunglasses, and the absolute difference between the two. This visualization demonstrates that the FaceLab app's modifications are localized exclusively to the sunglasses' region.}
\vspace{-4mm}
\label{fig:absolute}
\end{figure}

\begin{figure}[t]
    \centering
    \captionsetup[subfigure]{labelformat=empty}
    \begin{subfigure}[b]{1\linewidth}
        \begin{subfigure}[b]{1\linewidth}
            \centering
            \begin{subfigure}[b]{\linewidth}
                \begin{subfigure}[b]{0.28\linewidth}
                \caption{Sunglasses + blur Probe}
                    \includegraphics[width=\linewidth]{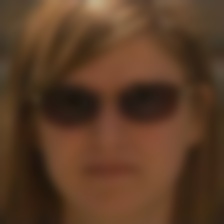}
                \end{subfigure}
                \begin{subfigure}[b]{0.28\linewidth}
                \caption{Mated Match Gallery}
                    \includegraphics[width=\linewidth]{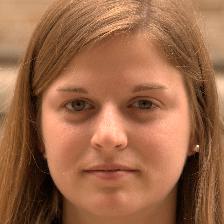}
                \end{subfigure}
                \begin{subfigure}[b]{0.28\linewidth}
                \caption{Non-Mated Match Gallery}
                    \includegraphics[width=\linewidth]{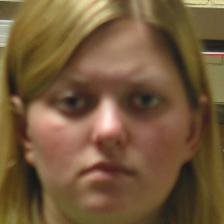}
                \end{subfigure}     

                 \begin{subfigure}[b]{0.28\linewidth}
                    \includegraphics[width=\linewidth]{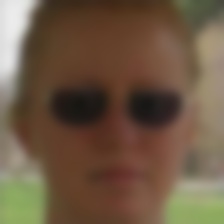}
                \end{subfigure}
                \begin{subfigure}[b]{0.28\linewidth}
                    \includegraphics[width=\linewidth]{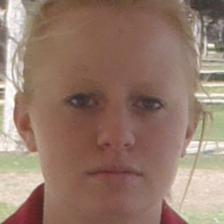}
                \end{subfigure}
                \begin{subfigure}[b]{0.28\linewidth}
                    \includegraphics[width=\linewidth]{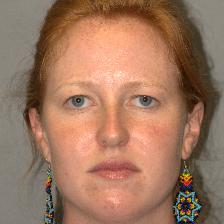}
                \end{subfigure}     
                
            \caption{(a) Caucasian Females}
            \end{subfigure}            
            
            \begin{subfigure}[b]{\linewidth}
            
                \begin{subfigure}[b]{0.28\linewidth}
                    \includegraphics[width=\linewidth]{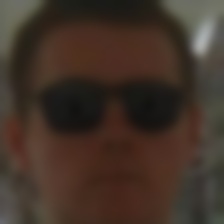}
                \end{subfigure}
                \begin{subfigure}[b]{0.28\linewidth}
                    \includegraphics[width=\linewidth]{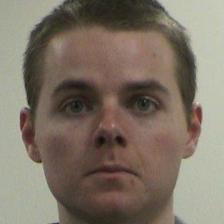}
                \end{subfigure}
                \begin{subfigure}[b]{0.28\linewidth}
                    \includegraphics[width=\linewidth]{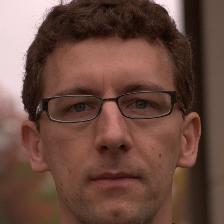}
                \end{subfigure}

                \begin{subfigure}[b]{0.28\linewidth}
                    \includegraphics[width=\linewidth]{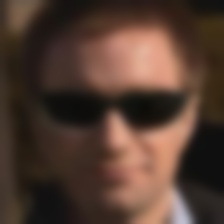}
                \end{subfigure}
                \begin{subfigure}[b]{0.28\linewidth}
                    \includegraphics[width=\linewidth]{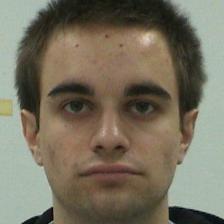}
                \end{subfigure}
                \begin{subfigure}[b]{0.28\linewidth}
                    \includegraphics[width=\linewidth]{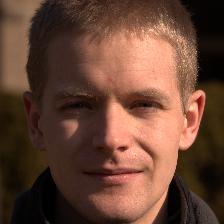}
                \end{subfigure}
              
                \caption{(b) Caucasian Males}
            \end{subfigure}
        \end{subfigure}
        
    \end{subfigure}
   \caption{This figure shows the probe with sunglasses + blur, the gallery image that is the highest similarity of the right / same identity, and the gallery image that matched to it as the rank-one similarity but the wrong / different person.}
   \vspace{-4mm}
\label{fig:sungl_blur_example}
\end{figure}

\begin{figure}[t]
    \centering
    \begin{subfigure}[b]{1\linewidth}
        \includegraphics[width=\linewidth]{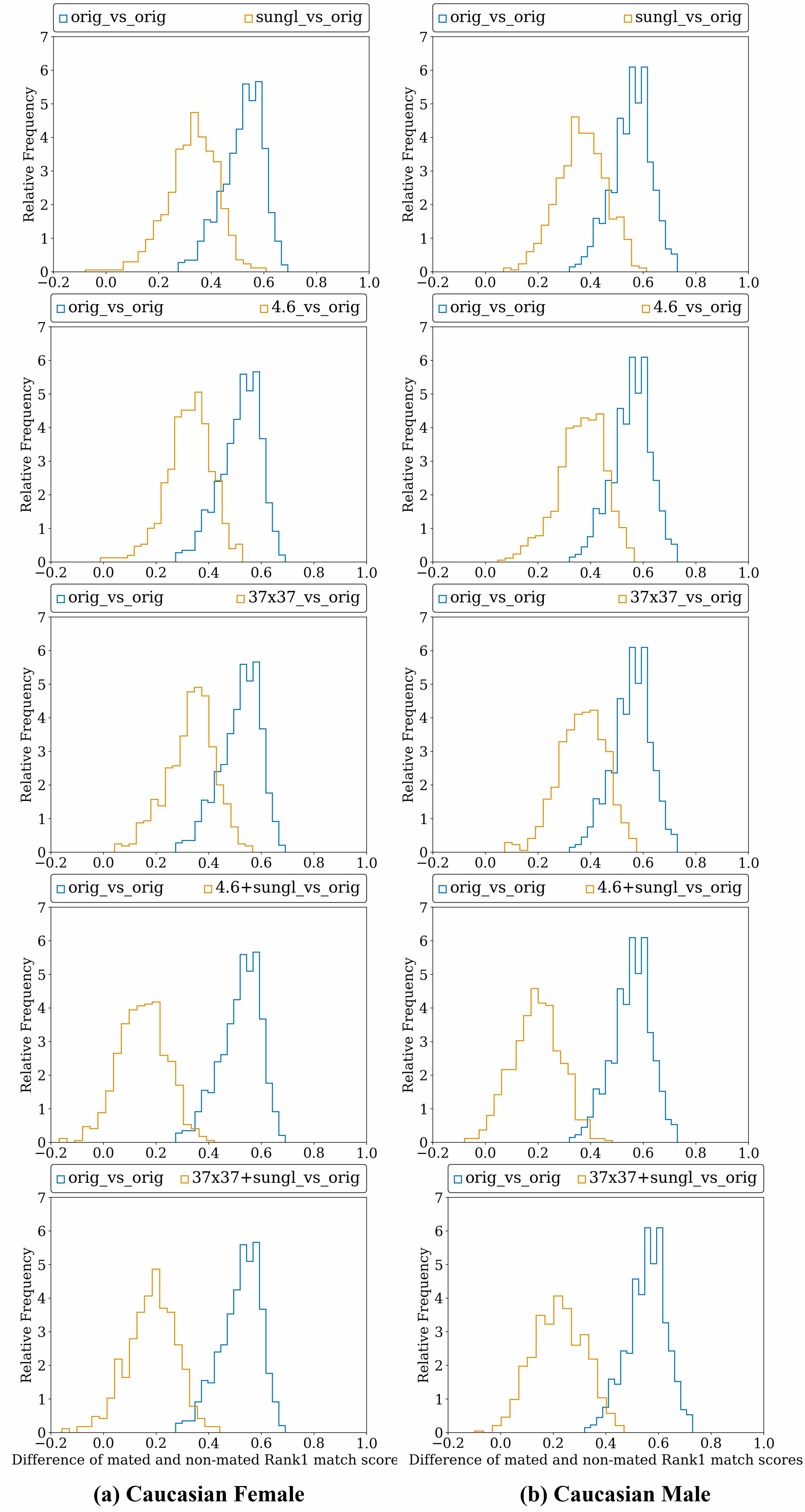}
    \end{subfigure}
    \vspace{-5mm}
   \caption{Comparison mated and non-mated distributions for original probe images and different conditions of degraded probes (top to bottom): 1) sunglasses, 2) blur of $\sigma = 4.6$, 3) 37x37 resolution, 4) sunglasses + blur, 5) sunglasses + 37x37 resolution. The face matcher is AdaFace.}
\vspace{-4mm}
\label{fig:ada_simi_prob}
\end{figure}

\begin{figure}[t]
    \centering
    \begin{subfigure}[b]{1\linewidth}
        \includegraphics[width=\linewidth]{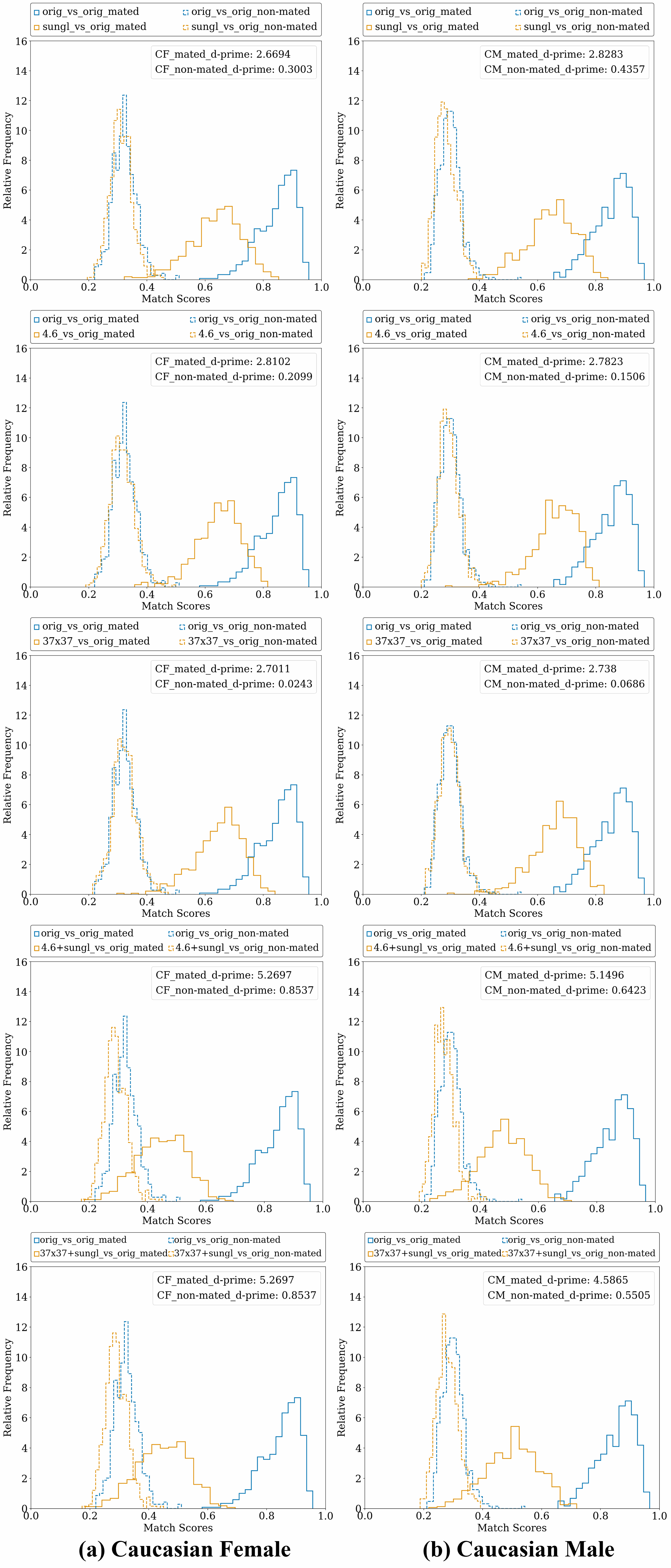}
    \end{subfigure}
   \caption{Comparison mated and non-mated distributions for original probe images and different conditions of degraded probes (top to bottom): 1) sunglasses, 2) blur of $\sigma = 4.6$, 3) 37x37 resolution, 4) sunglasses + blur, 5) sunglasses + 37x37 resolution. The face matcher is ArcFace.}
\vspace{-4mm}
\label{fig:sup_simi_prob}
\end{figure}

\begin{figure}[t]
    \centering
    \begin{subfigure}[b]{1\linewidth}
        \includegraphics[width=\linewidth]{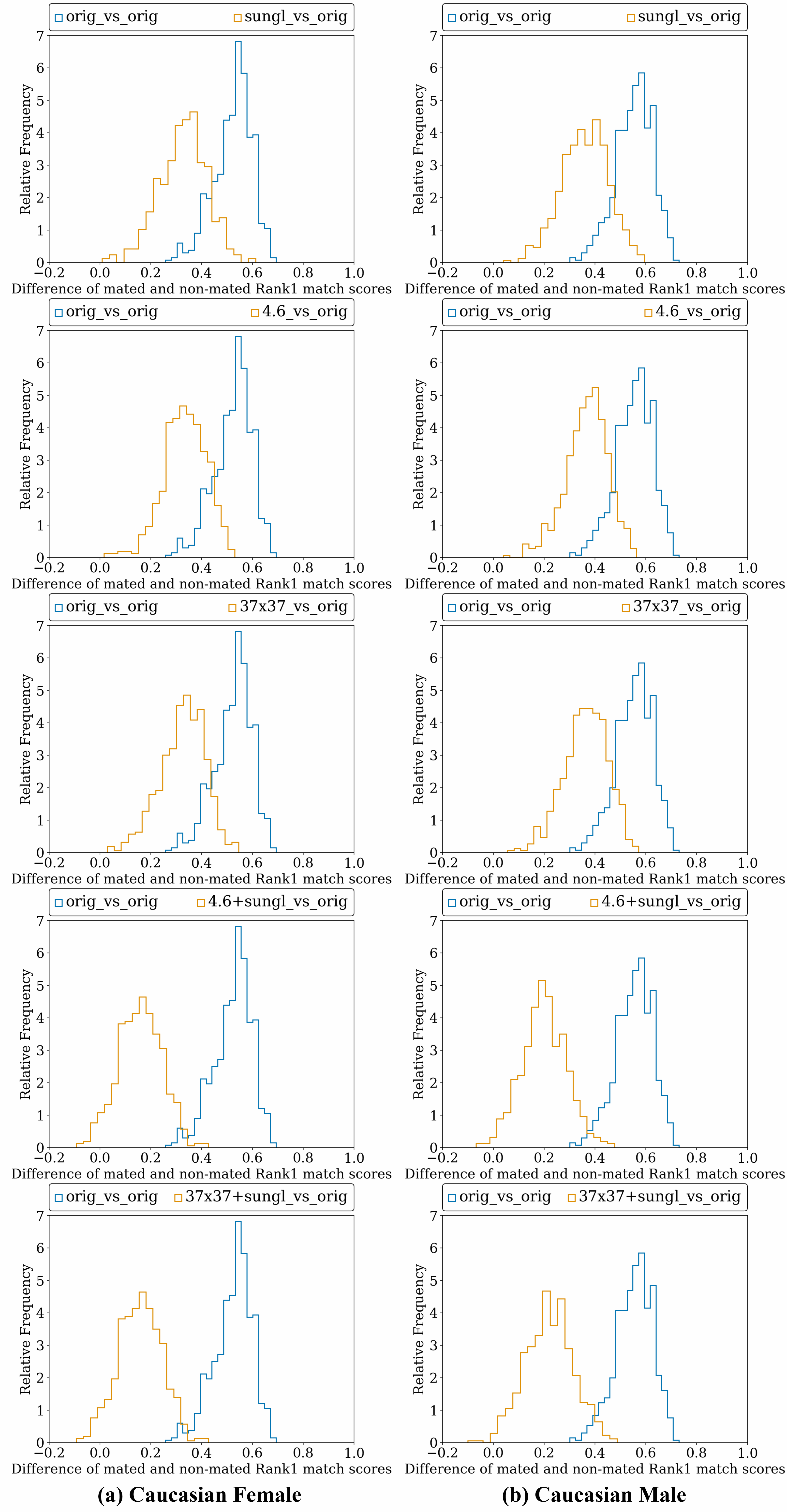}
    \end{subfigure}
   \caption{Score difference (mated - non-mated) distributions for original probe images and different conditions of degraded probes (top to bottom): 1) sunglasses, 2) blur of $\sigma = 4.6$, 3) 37x37 resolution, 4) sunglasses + blur, 5) sunglasses + 37x37 resolution. The face matcher is ArcFace.}
\vspace{-4mm}
\label{fig:sup_diff_prob}
\end{figure}

\begin{figure*}[t]
    \centering
    \begin{subfigure}[b]{1\linewidth}
        \includegraphics[width=\linewidth]{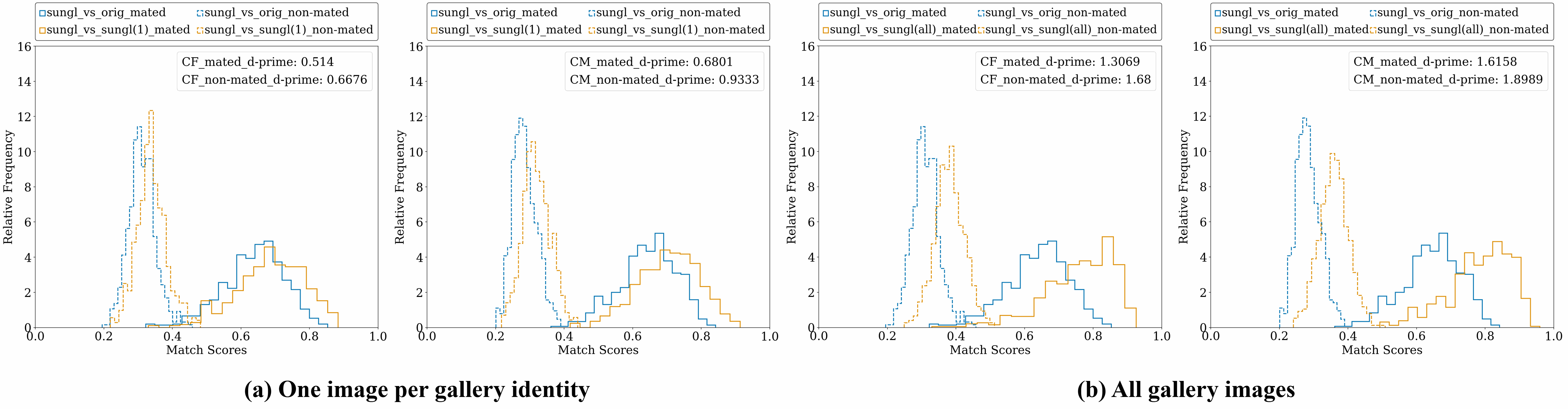}
    \end{subfigure}
    \vspace{-5mm}
   \caption{Comparing mated and non-mated distributions for sunglasses-added probe images under different gallery degradation conditions: (a) sunglasses randomly added to one image per identity in the gallery, (b) sunglasses added to all gallery images. Distributions are shown separately for male and female subjects. It used the ArcFace recognition matcher.}

\label{fig:sup_simi_solu}
\end{figure*}

\begin{figure*}[t]
    \centering
    \begin{subfigure}[b]{1\linewidth}
        \includegraphics[width=\linewidth]{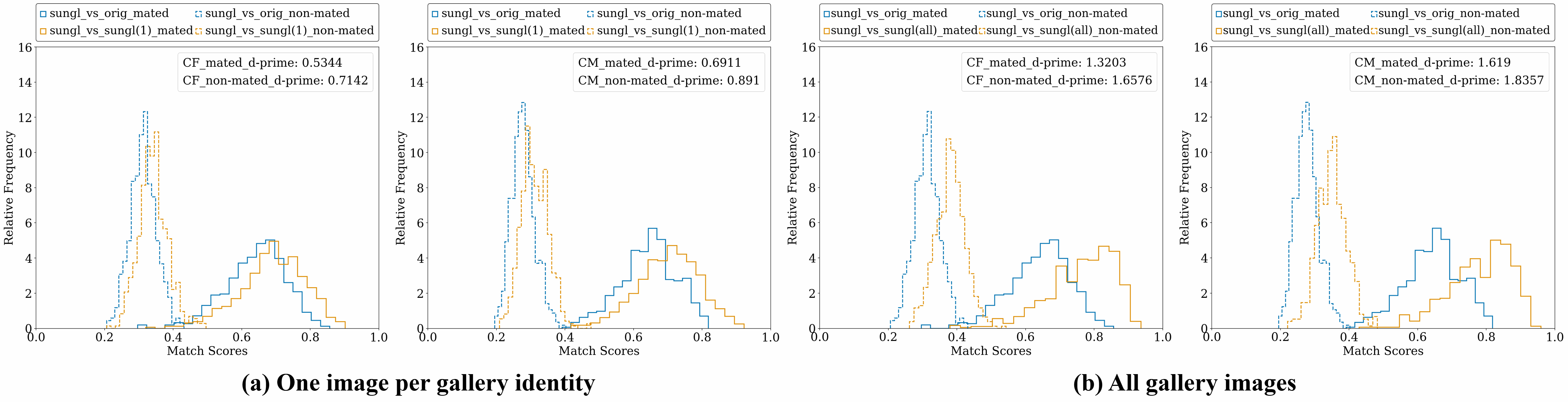}
    \end{subfigure}
    \vspace{-5mm}
   \caption{Comparing mated and non-mated distributions for sunglasses-added probe images under different gallery degradation conditions: (a) sunglasses randomly added to one image per identity in the gallery, (b) sunglasses added to all gallery images. Distributions are shown separately for male and female subjects. It used the AdaFace recognition matcher.}
\label{fig:ada_simi_solu}
\end{figure*}

\begin{figure}[t]
    \centering
    \begin{subfigure}[b]{1\linewidth}
        \includegraphics[width=\linewidth]{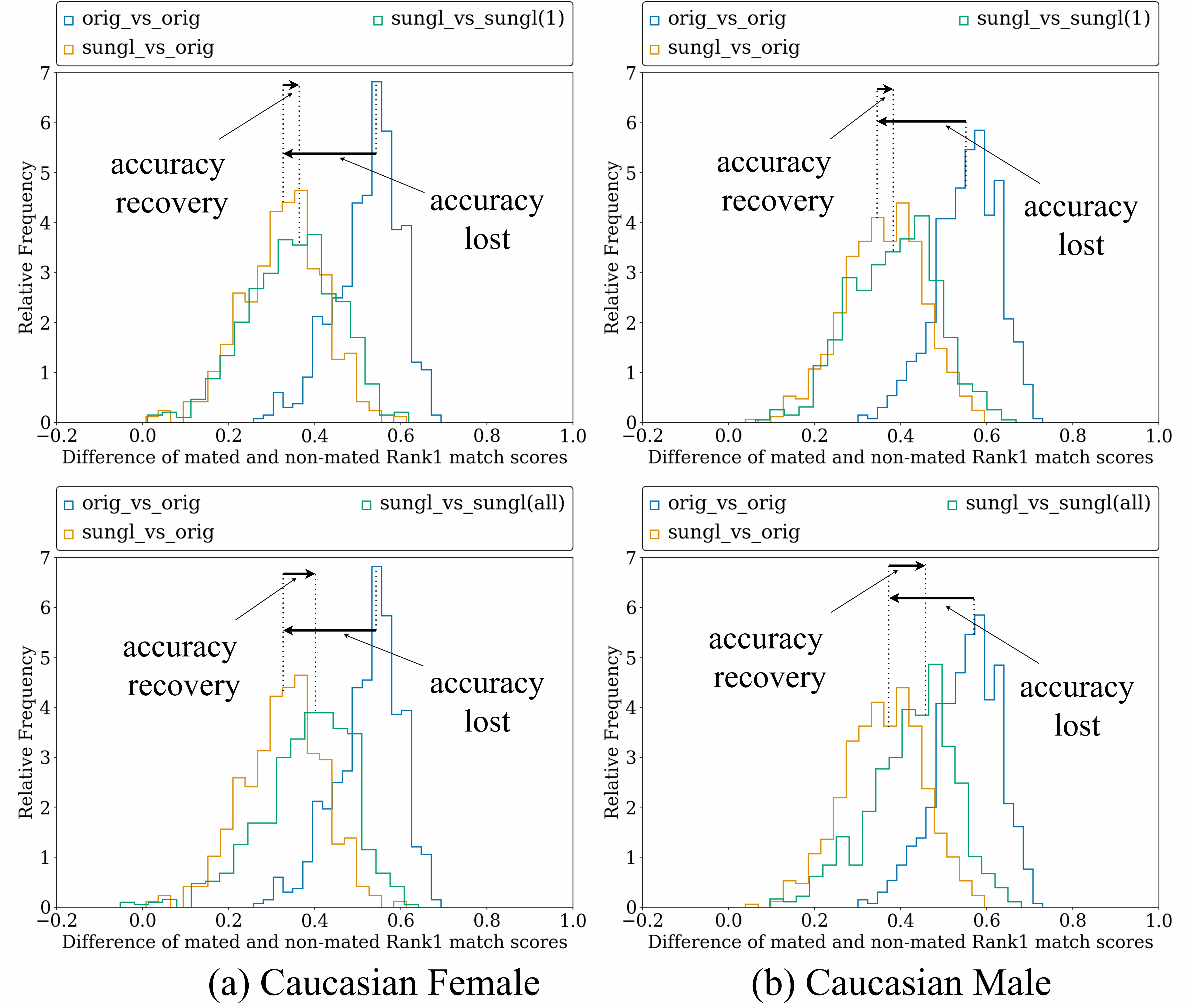}
    \end{subfigure}
    \vspace{-5mm}
   \caption{Similarity difference (mated - non-mated) distributions. Comparing with the \textcolor[HTML]{0c80d5}{baseline}, adding sunglasses to probe images causes a large \textit{accuracy loss}. To counter this effect, we add sunglasses to one (all) image(s) per identity in the gallery to achieve good \textit{accuracy recovery}, as shown in Top (Bottom) row. It used the ArcFace recognition matcher.}
\vspace{-4mm}
\label{fig:sup_diff_solu}
\end{figure}

\thispagestyle{empty}
The experiments in the paper primarily utilize the AdaFace  matcher. This supplementary material presents the same experimental results conducted with the ArcFace recognition matcher, thereby validating the conclusions of the paper. Specifically, this supplementary material provides:

\begin{itemize}

    \item Figure~\ref{fig:diff_types} shows some examples of different types of synthetic sunglasses added by the FaceLab app. This app will add different types of realistic sunglasses based on the images.

    \item Figure~\ref{fig:images/sup_small_experiment} demonstrates the high identity-preservation of sunglasses augmentation produced by the FaceLab app. The minimal differences and high similarity between the original and augmented images validate the use of FaceLab app for sunglasses addition in our research.

    \item Figure~\ref{fig:absolute} shows the FaceLab app’s modifications are localized exclusively to the sunglasses’ region.

    \item Figure~\ref{fig:sungl_blur_example} shows shows the probe with sunglasses + blur, the gallery image that is the highest similarity of the right identity, and the gallery image that matched to it as the rank-one similarity but the wrong person.

    \item Figure~\ref{fig:ada_simi_prob} shows a comparison of mated and non-mated distributions for original probe images and various conditions of degraded probes using the AdaFace  matcher.
   
    \item Figure~\ref{fig:sup_simi_prob} shows a comparison of mated and non-mated distributions for original probe images and various conditions of degraded probes using the ArcFace  matcher.
   
    \item Figure~\ref{fig:sup_diff_prob} shows score difference (mated - non-mated) distributions for original probe images and different conditions of degraded probes using the ArcFace  matcher.
   
    \item Figure~\ref{fig:sup_simi_solu} compares mated and non-mated distributions for sunglasses-added probe images under different gallery degradation conditions using the ArcFace  matcher.

    \item Figure~\ref{fig:ada_simi_solu} compares mated and non-mated distributions for sunglasses-added probe images under different gallery degradation conditions using the AdaFace  matcher.
    
    \item Figure~\ref{fig:sup_diff_solu} compares difference (mated - non-mated) distributions for sunglasses-added probe images under different gallery degradation conditions using the ArcFace  matcher.
    
    \item Table~\ref{tab:sup_wass_prob} shows the Wasserstein distances of the differences in (mated - non-mated) distributions in Figure~\ref{fig:sup_simi_prob}, under various effects to the probe images.

\end{itemize}

\newpage
\setlength{\tabcolsep}{1.2mm}

\begin{table}[]
\centering
\begin{tabular}{c|c|c}
\hline
 Effects added to probe images                                                     & Demographic & Probe vs. original gallery \\ \hline
\multirow{2}{*}{Sunglasses}                            & \multicolumn{1}{c|}{CF}                  & 0.197                 \\  
                                                                                    & \multicolumn{1}{c|}{CM}                  & 0.194                 \\ \hline
\multirow{2}{*}{Gaussian blur ($\sigma = 4.6$)}                 & \multicolumn{1}{c|}{CF}                  & 0.190                 \\ 
                                                                                    & \multicolumn{1}{c|}{CM}                  & 0.187                  \\ \hline
\multirow{2}{*}{Low-resolution (37x37)}                & \multicolumn{1}{c|}{CF}                  & 0.199                 \\  
                                                                                    & \multicolumn{1}{c|}{CM}                  & 0.191                 \\ \hline
\multirow{2}{*}{Sunglasses and Gaussian blur}  & \multicolumn{1}{c|}{CF}                  & 0.366                 \\  
                                                                                    & \multicolumn{1}{c|}{CM}                  & 0.354                 \\ \hline
\multirow{2}{*}{Sunglasses and low-resolution} & \multicolumn{1}{c|}{CF}                  & 0.366                 \\  
                                                                                    & \multicolumn{1}{c|}{CM}                  & 0.334                 \\ \hline
\end{tabular}
\caption{Wasserstein distances of the differences in (mated - non-mated) distributions in Figure~\ref{fig:sup_simi_prob}, under various effects to the probe images.}
\vspace{-2mm}
\label{tab:sup_wass_prob}
\end{table}

\end{document}